\documentclass{article}

\usepackage[preprint]{neurips_2025}

\usepackage[utf8]{inputenc} 
\usepackage[T1]{fontenc}    
\usepackage{hyperref}       
\usepackage{url}            
\usepackage{booktabs}       
\usepackage{amsfonts}       
\usepackage{nicefrac}       
\usepackage{microtype}      
\usepackage{xcolor}         

\usepackage{graphicx}
\usepackage{color}
\usepackage{colortbl}
\usepackage{multirow}
\usepackage{wrapfig}
\usepackage{xspace}
\usepackage{pifont}
\usepackage{amsmath} 
\usepackage{amssymb}
\usepackage{subfig}
\usepackage{float}

\makeatletter
\newcommand{\thickhline}{
    \noalign {\ifnum 0=`}\fi \hrule height 1pt
    \futurelet \reserved@a \@xhline
}
\newcommand{\ie}{\emph{i.e.}\xspace}
\newcommand{\eg}{\emph{e.g.}\xspace}
\newcommand{\etc}{\emph{etc.}\xspace}

\newcommand*\samethanks[1][\value{footnote}]{\footnotemark[#1]}

\title{\raisebox{-0.01\textheight}{\includegraphics[width=0.06\textwidth]{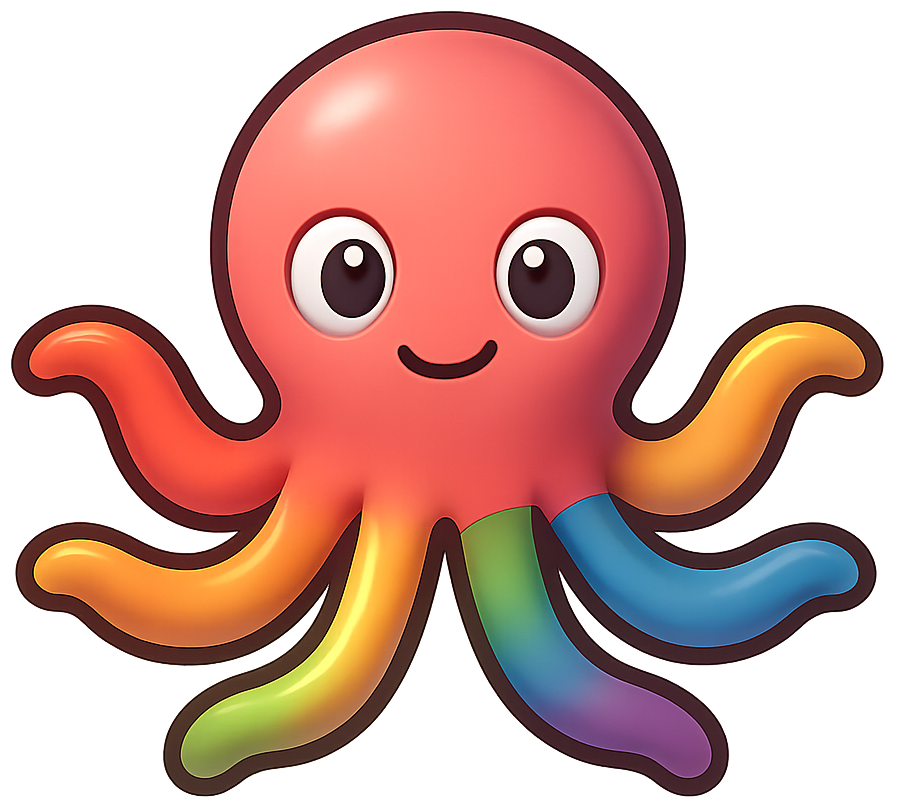}} OctoNav: Towards Generalist Embodied Navigation}

\author{%
  Chen Gao$^{1, 2}$\thanks{Equal contribution}\qquad
  Liankai Jin$^{1}$\samethanks\qquad
  Xingyu Peng$^{1, 4}$\samethanks\qquad
  Jiazhao Zhang$^{3}$\\
  \textbf{Yue Deng$^{1, 4}$\qquad
  Annan Li$^{1}$\qquad
  He Wang$^{3}$\qquad
  Si Liu$^{1}$\thanks{Corresponding author}}\\
  \textsuperscript{\textnormal{1}}Beihang University\qquad \textsuperscript{\textnormal{2}}National University of Singapore\\ \textsuperscript{\textnormal{3}}Peking University\qquad
  \textsuperscript{\textnormal{4}}Zhongguancun Academy\\
  \\
  \small{Project page: \url{https://buaa-colalab.github.io/OctoNav}}
}

\begin{document}

\maketitle

\begin{figure}[ht]
    \begin{center}
    \vspace{-6mm}
    \centerline{\includegraphics[width=1\linewidth]{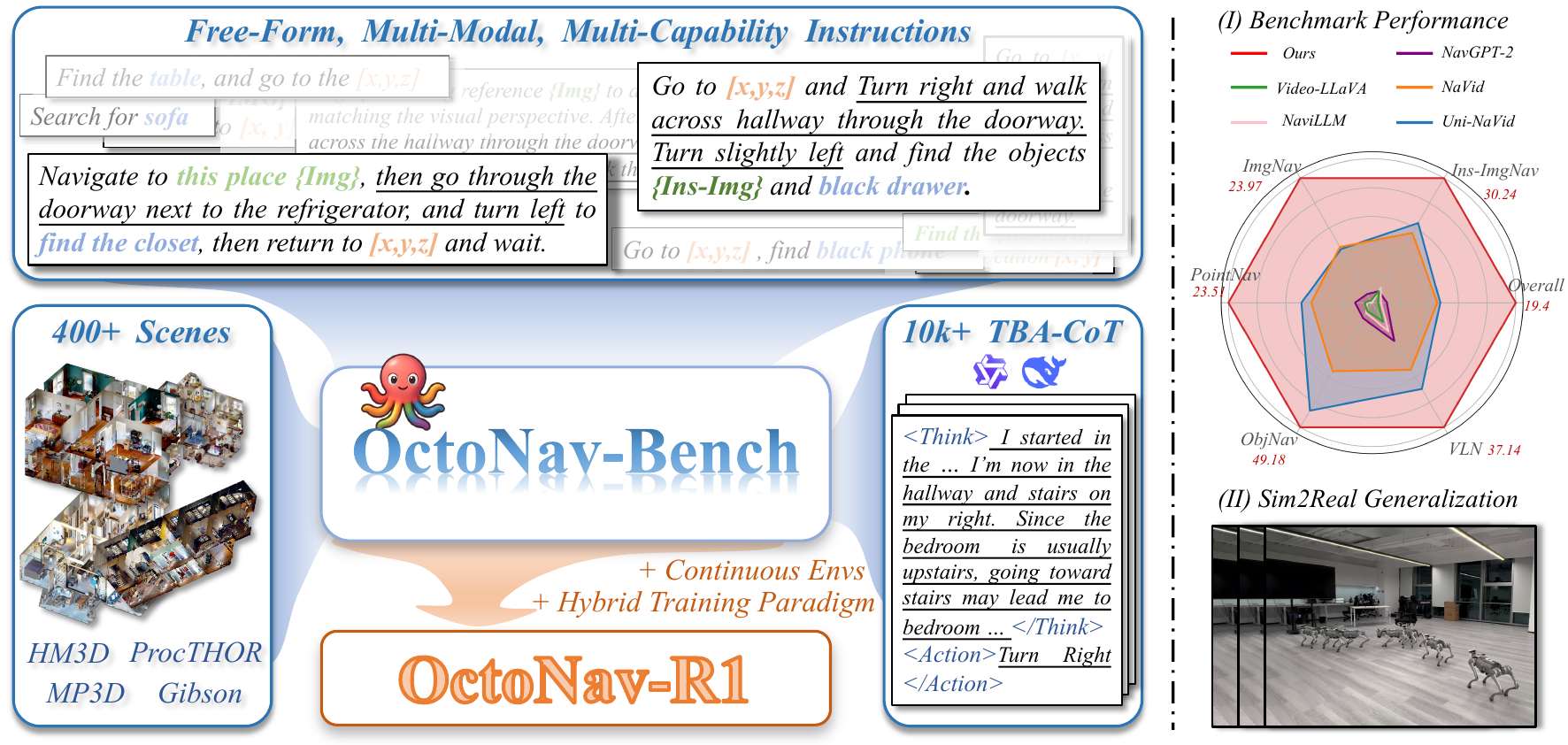}}
        \caption{
On the left, we present the large-scale OctoNav-Bench, which contains diverse instruction-trajectory pairs and the elaborate TBA-CoT dataset across numerous scenes. 
Based on OctoNav-Bench and our method/training designs, we introduce a VLA-based method, termed OctoNav-R1. 
On the right, \emph{(I)} demonstrates the performance comparisons on OctoNav-Bench, where we provide a fine-grained breakdown of accuracy across various navigation capabilities. OctoNav-R1 outperforms previous methods in all capabilities, demonstrating its versatility. 
\emph{(II)} presents a robot demo in the real world, which is driven by the OctoNav-R1, showing its preliminary sim2real generalization.
        }
        \label{fig:teaser}
        \vspace{-5mm}
    \end{center}
\end{figure}

\begin{abstract}
Embodied navigation stands as a foundation pillar within the broader pursuit of embodied intelligence. However, previous navigation research is divided into different tasks/capabilities, \eg, ObjNav, ImgNav and VLN, where they differ in task settings/objectives and modalities, making datasets and methods are designed individually. 
In this work, we take steps toward generalist navigation agents, which can follow free-form instructions that include arbitrary compounds of multi-modal and multi-capability. 
To achieve this, we propose a large-scale benchmark and corresponding method, termed OctoNav-Bench and OctoNav-R1. 
Specifically, OctoNav-Bench features continuous environments and is constructed via a designed automatic annotation pipeline. We thoroughly craft instruction-trajectory pairs for imitation learning, where instructions are diverse in free-form with arbitrary modality and capability. Also, we elaborately construct a Think-Before-Action (TBA-CoT) dataset within OctoNav-Bench to provide the thinking process behind actions.
For OctoNav-R1, we build it upon MLLMs and adapt it to a VLA-type model, which can produce low-level actions solely based on 2D visual observations. 
Moreover, we design a Hybrid Training Paradigm (HTP) that consists of three stages, \ie, Action-/TBA-SFT, Nav-GPRO, and Online RL stages. Each stage contains specifically designed learning policies and rewards. 
Importantly, for TBA-SFT and Nav-GRPO designs, we are inspired by the OpenAI-o1 and DeepSeek-R1, which show impressive reasoning ability via thinking-before-answer. Thus, we aim to investigate how to achieve thinking-before-action in the embodied navigation field, to improve model's reasoning ability toward generalists. 
Specifically, we propose TBA-SFT to utilize the TBA-CoT dataset to fine-tune the model as a cold-start phrase and then leverage Nav-GPRO to improve its thinking ability. 
Finally, OctoNav-R1 shows superior performance compared with the previous methods.
\end{abstract}
\vspace{-1mm}
\section{Introduction}
\label{sec:introduction}

\vspace{-0.5mm}
Embodied navigation aims to enable agents to perceive, reason, and move within previously-unseen environments to reach specified goals, constituting a crucial research direction within embodied intelligence.
However, current embodied navigation is fragmented into narrowly defined tasks such as object goal (ObjNav)~\cite{chaplot2020object,gao2023room}, point goal (PointNav)~\cite{habitat2020sim2real,zhao2021surprising,partsey2022mapping}, image goal (ImgNav)~\cite{zhu2017target,sun2023fgprompt}, instance-image goal (Ins-ImgNav)~\cite{habitatchallenge2023,krantz2023navigating,krantz2022instance}, and vision-language navigation (VLN)~\cite{anderson2018vision,krantz2020beyond,qi2020reverie}, where \emph{they differ in task settings, input modalities, and objectives} \etc 
For instance, PointNav requires agents to reach target coordinates. ImgNav and Ins-ImgNav require agents to find a scene or object that match given reference images.
Therefore, separated benchmarks and methods are proposed for each task, and
their rigid separation poses a challenge for building generalist navigation agents. 
Agents trained on specific task definitions lack the flexibility to handle diverse tasks, restricting models to specific capabilities.
For example, a VLN agent can not achieve visual goals described by reference images.

Ideally, a generalist agent is expected to follow free-form instruction, \ie, \emph{beyond isolated task-specific instruction, also compound of multi-modal and multi-task/capability instructions.}
For example, as shown in Fig.~\ref{fig:teaser}, instructions like ``navigate to this place \{Image\}, then go through the doorway next to the refrigerator, and turn left to find the closet, then return to \{x,y,z\} and wait'', span coordinate, visual, language modalities and PointNav, ImgNav, VLN tasks/capabilities. 
Thus, we aspire to move beyond narrow task specialization toward generalist, \ie, investigating the construction of generalist navigation agents capable of following free-form, multi-modal, and multi-capability instructions.

Most recent works take preliminary attempts for building generalizable navigation agents, but key limitations remain. 
For example, 
GOAT-Bench~\cite{khanna2024goat} and LHPR-VLN~\cite{song2024towards} benchmarks are built for lifelong and long-horizon navigation learning, respectively. 
However, as shown in Tab.~\ref{tab:benchmark}, these benchmarks only contain two capabilities. Moreover, they are not free-form instructions, meaning each instruction only involves a single capability or modality. Thus, they are essentially multi-task benchmarks with various tasks being independent, failing to support generalist navigation agents.

In this work, we introduce \textbf{OctoNav-Bench}, a large-scale and unified benchmark specifically designed for generalist embodied navigation, which is distinguished by the following core features.
\textbf{\emph{(1) Large-scale Annotations:}} OctoNav-Bench encompasses $400+$ diverse 3D scenes sourced from widely used HM3D and Gibson \etc
Also, OctoNav-Bench provides $45k+$ annotated instruction-trajectory pairs via the designed automatic annotation pipeline, supporting large-scale training.
\textbf{\emph{(2) Free-form, Multi-Model and Multi-capability Instructions:}} The instructions are generated in free-form descriptions. First, the capabilities included in the instruction are sampled from arbitrary combinations of ObjNav, PointNav, ImgNav, Ins-ImgNav, and VLN, \ie, each instruction contains multiple navigation capabilities simultaneously. Moreover, these instructions are multimodal, incorporating textual, visual (\eg, reference scene-/object-level images), and spatial (\eg, coordinates) descriptions. 
\textbf{\emph{(3) TBA-CoT Dataset:}} We leverage Qwen-VL and DeepSeek-R1 to construct a Think-Before-Action Chain-of-Thought (TBA-CoT) dataset, which captures the deliberative reasoning process behind each action decision. Such a dataset can be used to supervise and enhance the agent's reasoning ability.
\textbf{\emph{(4) Continuous Environments with RL Support:}} Unlike discrete or graph-based settings~\cite{anderson2018vision,qi2020reverie}, OctoNav-Bench provides continuous simulation environments, allowing agents to move freely and acquire visual observations at arbitrary locations. Thus, it supports active learning like online RL.

In addition to the benchmark, we further propose \textbf{OctoNav-R1}, a VLA-based model designed and trained on OctoNav-Bench, and is distinguished by the following key aspects:
\textbf{\emph{(1) Free-form, Multi-modal and Multi-capability Instruction Following:}} OctoNav-R1 can accept free-form instructions that comprise multi-modal and multi-capability. Based on step-wise egocentric visual observations, the model can directly generate a sequence of low-level actions (\eg, move forward, turn left/right), enabling it to follow complex instructions in a unified manner.
\textbf{\emph{(2) RL-enhanced VLA Hybrid Training Paradigm:}} Unlike conventional VLA models that are typically fine-tuned via SFT on static datasets, OctoNav-R1 are trained by the proposed Hybrid Training Paradigm (HTP). Specifically, we integrate RL into the VLA training pipeline, making HTP combine Action-/TBA-SFT, Nav-GRPO, and online RL stages. 
\textbf{\emph{(3) Thinking-Before-Action:}} Inspired by the long CoT reasoning within DeepSeek-R1~\cite{deepseekai2025deepseekr1incentivizingreasoningcapability}, we argue that previous VLA models, which directly map observations to actions, lack explicit thinking processes and struggle with complicated tasks.
Therefore, we leverage the TBA-CoT dataset to train OctoNav-R1 via TBA-SFT and Nav-GRPO, endowing the model with the ability to jointly produce thinking thoughts and action sequences.
\textbf{\emph{(4) Initial Sim2Real Generalization:}} We deploy OctoNav-R1 on physical robots, and observe preliminary sim-to-real transfer ability without real-world fine-tuning. It further confirms the annotated OctoNav-Bench and designed OctoNav-R1.

\section{Related Work}
\label{sec:related}
\vspace{-1mm}

\noindent\textbf{Large Model for Embodied Navigation.}
Recent works~\cite{long2024instructnav,lin2025navcot,bao2025enhancing,shi2025llm,zhang2025flexvln,yin2025unigoal,cheng2024navila} explored the integration of LLMs and MLLMs into agents. 
Early approaches PaLM-E~\cite{Driess2023PaLMEAE} incorporate multi-modal tokens, enabling high-level planning for manipulation and navigation. RT-2~\cite{zitkovich2023rt} extends this idea by directly predicting low-level actions for closed-loop control. 
RoboFlamingo~\cite{li2023vision} leverages vision-language pretraining to generate policies, emphasizing efficient adaptation. 
NaVid~\cite{zhang2024navid} fine-tunes a video-based MLLMs, achieving more promising results on specific navigation tasks. 
Other navigation agents like NavGPT~\cite{zhou2024navgpt,zhou2024navgpt2}, MapGPT~\cite{Chen2024MapGPTMP}, and TopV-Nav~\cite{zhong2024topv} treat LLMs/MLLMs as zero-shot agents, relying on prompt learning.
Besides, some works~\cite{zhang2025uninavid,zheng2024towards,wang2022towards} investigate unified models. 
For instance, Uni-NaVid~\cite{zhang2025uninavid} and NaviLLM~\cite{zheng2024towards} fine-tune MLLMs to improve task transfer ability. Note that NaviLLM only applies navigation in discrete environments. Moreover, they leverage existing datasets via direct multi-task learning. 
Therefore, despite these methods showing potential for generalization, they can only handle different tasks separately and fall short when faced with free-form instructions that simultaneously include multi-modal and multi-capability. 

\noindent\textbf{Reinforcement Learning for Large Model.}
Prior large model works~\cite{yao2023react,besta2024topologies,yao2023tree,long2023large,besta2024graph} show that stepwise reasoning, such as chain-of-thought (CoT)~\cite{wei2022chain}, can improve performance on complex tasks~\cite{li2024chain,zhang2024autoregressive+,feng2023towards}. 
Recently, reinforcement learning (RL) has emerged as a promising strategy~\cite{deepseekai2025deepseekr1incentivizingreasoningcapability,shao2024deepseekmath,lin2025cppo,ramesh2024group,li2025adaptive} for improving model reasoning ability. Models like DeepSeek-R1~\cite{deepseekai2025deepseekr1incentivizingreasoningcapability} demonstrate that even outcome-only rewards can guide LLMs to develop reasoning behaviors without step-level annotations.
More recently, the training paradigm of DeepSeek-R1 has been applied to the MLLMs literature, \eg, Vision-R1~\cite{huang2025vision}, Video-R1~\cite{feng2025video,zhang2025tinyllava} construct multi-modal CoT datasets. 
Despite these improvements, most efforts are limited to static images or constrained tasks.
There is currently no research that investigates how DeepSeek-R1-style training philosophy can be applied to robot scenarios. 
Considering the complexity of visual observations and free-form instructions, we build our OctoNav-Bench with a TBA-CoT dataset. By the devised HTP, our OctoNav-R1 gains strong planning capabilities for navigation.
\vspace{-1mm}
\section{OctoNav-Bench}
\label{sec:bench}
\vspace{-1mm}
\begin{figure}[t]
    \begin{center}
    \centerline{\includegraphics[width=1\linewidth]{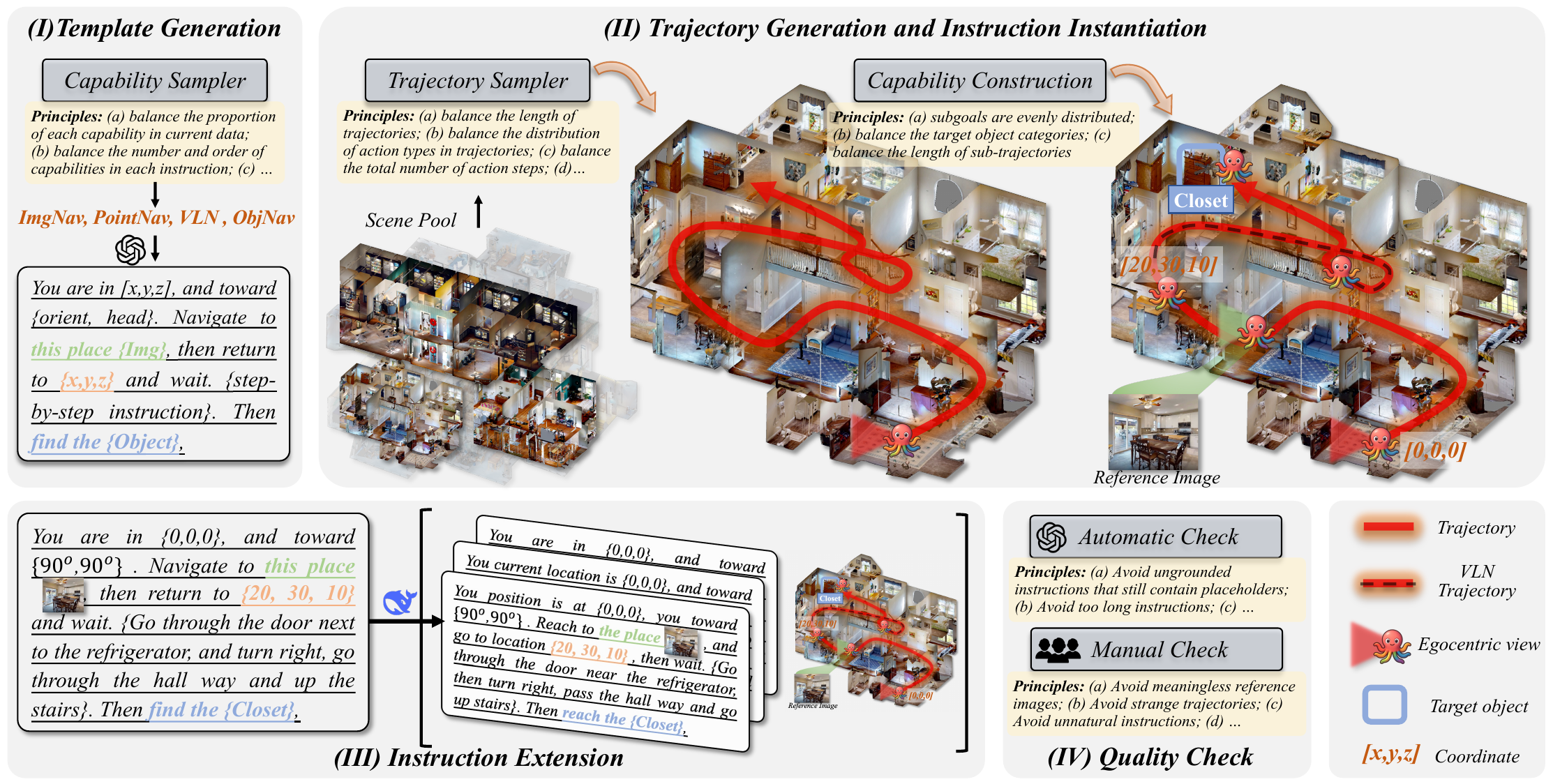}}
    \vspace{-2mm}
\caption{\textbf{The automatic construction pipeline of \emph{OctoNav-Bench.}}  \emph{(I) Template Generation.} We generate diverse instruction templates, where multiple capabilities are involved and specific elements are represented via placeholders. \emph{(II) Trajectory Generation and Instruction Instantiation.} We extract elements along the sampled trajectory and instantiate the instruction by grounding the placeholders with corresponding elements. \emph{(III) Instruction Extension.} We extend instructions with their variants.
\emph{(IV) Quality Check.} We apply automatic and manual verification stages.
Best viewed in color.}
    \vspace{-7mm}
        \label{fig:data}
    \end{center}
\end{figure}
We propose OctoNav-Bench, which is a large-scale benchmark for facilitating generalist embodied navigation. 
The scalable and automated pipeline is devised for benchmark building (shown in Fig.~\ref{fig:data}).

\vspace{-1mm}
\subsection{Instruction-Trajectory Pairs Construction}

\vspace{-0.5mm}
\noindent\textbf{Template Generation.}
In this stage, we aim to generate diverse instruction templates, where each instruction can incorporate multiple navigation capabilities simultaneously. 
As shown in Fig.~\ref{fig:data} (\uppercase\expandafter{\romannumeral1}), we employ a capability sampler to determine which capabilities are included in each instruction. 
The capability sampler follows predefined principles, \eg, balancing the proportion of each capability. 
Then, we leverage GPT to generate templates according to the involved capabilities. 
Currently, since task-specific elements (\eg, reference images for ImgNav) are not determined, we adopt placeholders to represent these elements, enabling flexible instantiation during the following stages.

\vspace{-0.5mm}
\noindent\textbf{Trajectory Generation and Instruction Instantiation.}
To construct \emph{instruction-trajectory} pairs, we generate diverse navigation trajectories and instantiate multi-modal instructions accordingly. 
Concretely, as shown in Fig.~\ref{fig:data}(II), we curate a scene pool consisting of over $400$ indoor environments drawn from MP3D~\cite{chang2017matterport3d}, HM3D~\cite{ramakrishnan2021hm3d}, Gibson~\cite{xia2018gibson}, and ProcTHOR~\cite{deitke2022️procthor}. A scene is first sampled from this pool, after which a trajectory is sampled within the selected environment using a customized trajectory sampler (appendix~\ref{append:bench-construction-traj}). To ensure that the sampled trajectories exhibit diversity and realism, we follow principles and apply constraints during sampling, \eg, balance trajectory lengths and action types.
Once a full trajectory is obtained, we instantiate the corresponding instruction to align the trajectory via the capability construction mechanism. Concretely, we extract visual and positional attributes along the trajectory to ground the placeholders in the instruction template (\eg, image, coordinate point, object category). Starting from the initial location, we select multiple waypoints along the trajectory as sub-goals, where each corresponds to a distinct capability. For example, as shown in Fig.~\ref{fig:data}(II) the image observed at the first sub-goal becomes the reference image for ImgNav, and an object category near the final sub-goal serves as the final target for ObjNav. 
To ensure coverage and diversity, we also enforce several additional criteria and principles during capability construction, \eg, the selected sub-goals are spatially distributed to avoid redundancy
Through this process, we obtain a collection of grounded instruction-trajectory pairs.

\vspace{-0.5mm}
Finally, we conduct instruction extension and quality check stages to improve data diversity and quality. Overall, we adopt LLMs to expand each instruction into multiple semantically equivalent versions. We apply automatic and manual checks to filter and revise data. More details in appendix~\ref{append:bench-construction-check}.

\vspace{-1mm}
\subsection{Multi-Modal TBA-CoT Dataset Construction}
\vspace{-0.5mm}

\begin{figure}[t]
    \begin{center}
\centerline{\includegraphics[width=0.95\linewidth]{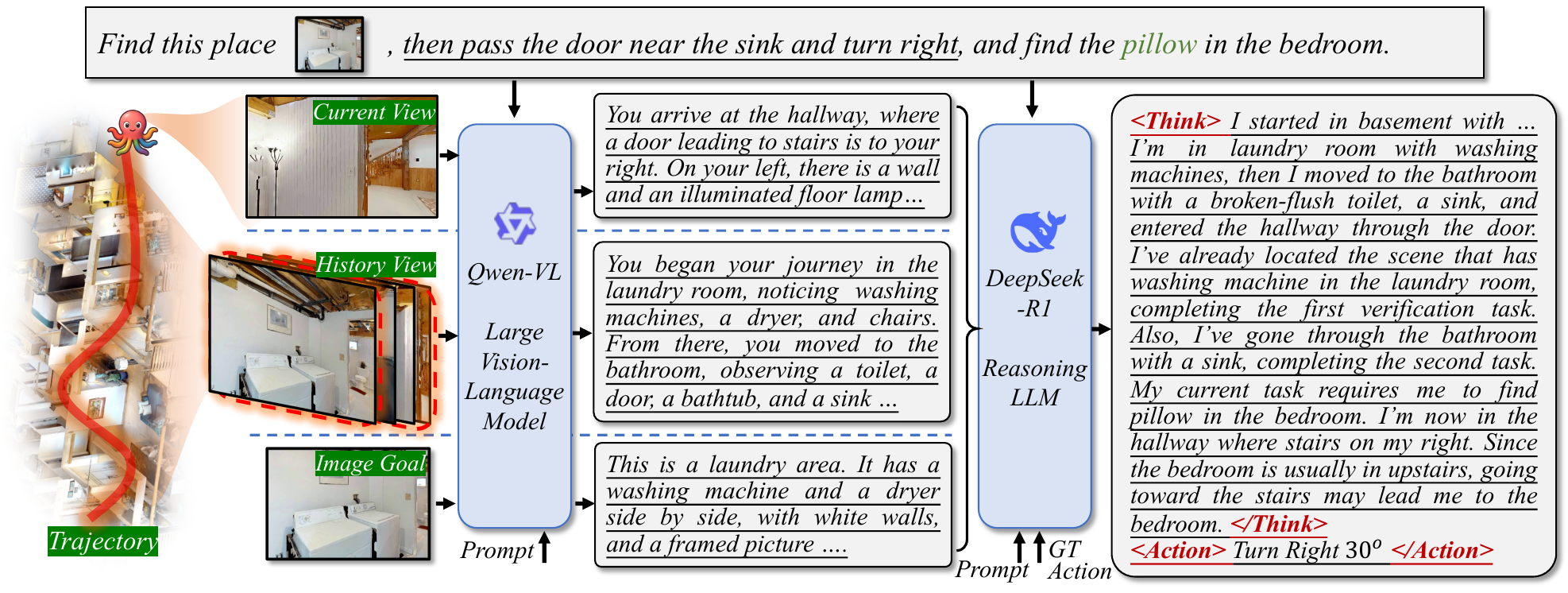}}
    \vspace{-2mm}
        \caption{\textbf{The automatic construction method of \emph{TBA-CoT.}} For the trajectories in OctoNav-Bench, we leverage Qwen-VL and DeepSeek-R1 to produce the thinking thoughts behind the action steps.}
    \vspace{-6mm}
        \label{fig:cot}
    \end{center}
\end{figure}

We design an automatic construction method for building the Think-Before-Action Chain-of-Thought (TBA-CoT) dataset, which is shown in Fig.~\ref{fig:cot}. Since OctoNav-Bench already contains \emph{instruction-trajectory} pairs, we aim to annotate the pseudo reasoning-thoughts behind actions along the trajectory.
Technically, at time step $t$, the agent has a ground-truth action \eg, \emph{turn right} $30^\circ$. We first extract its current view image, history view images, and reference images of ImgNav/Ins-ImgNav target (if any in the instruction).
For the current view, we prompt Qwen-VL to obtain an image description by feeding the current view image, the instruction, and structured prompts. Similarly, we summarize the historical views and reference image by inputting the instruction, corresponding images, and prompts.
Such a process converts the visual modality into language modality with key perceptual cues retention. 
We then aggregate and restructure these descriptions with appropriate prompts and feed them into DeepSeek-R1. Note that we also feed the ground-truth action of the current step, making LLM generate a detailed reasoning trace according to the reference action.

\begin{table}
\centering
\vspace{-1mm}
\caption{\textbf{Comparisons between \emph{OctoNav-Bench} and previous benchmarks}. $N_{T}$ denotes the task number. 
\emph{Mixed} indicates whether a single instruction integrates multiple capabilities.
\emph{Modality} is the modality within instructions, where
\emph{[V,L,P]} denote \emph{[vision, language, point]}.
\emph{TBA} presents the think-before-action annotations. 
\emph{DE, CE} denote the discrete and continuous environments.}
\resizebox{\textwidth}{!}{
{
\renewcommand{\arraystretch}{1.3}
\begin{tabular}{r||c||ccccccc||c||ccccc||c}
\hline \thickhline

 \rowcolor[HTML]{f8f9fa} \multicolumn{1}{c||}{}& & \multicolumn{7}{c||}{Instruction Capability} & \multicolumn{1}{c||}{} & \multicolumn{5}{c||}{Scenes} &  \\
\rowcolor[HTML]{f8f9fa} \multicolumn{1}{c||}{\multirow{-2}{*}{Benchmarks}} & {\multirow{-2}{*}{$N_{T}$}} & ObjNav & PointNav & VLN & ImgNav & Ins-ImgNav & Mixed & Modality & {\multirow{-2}{*}{TBA}} & MP3D & HM3D& Gibson & ProcTHOR& $N_{S}$&  {\multirow{-2}{*}{Env}}\\
\hline
\hline
R2R\cite{anderson2018vision}  & 22k & \ding{55} & \ding{55} & \ding{51} & \ding{55} & \ding{55} & \ding{55} & \emph{L} & \ding{55} & \ding{51} & \ding{55} & \ding{55} & \ding{55} & 90 & DE\\
R2R-CE~\cite{krantz2020beyond}  & 4.5k & \ding{55} & \ding{55} & \ding{51} & \ding{55} & \ding{55} & \ding{55} & \emph{L} & \ding{55} & \ding{51} & \ding{55} & \ding{55} & \ding{55} & 90 & CE\\
RxR-CE~\cite{krantz2020beyond}  & - & \ding{55} & \ding{55} & \ding{51} & \ding{55} & \ding{55} & \ding{55} & \emph{L} & \ding{55} & \ding{51} & \ding{55} & \ding{55} & \ding{55} & 90 & CE\\
SOON~\cite{zhu2021soon}  & 30k & \ding{51} & \ding{55} & \ding{55} & \ding{55} & \ding{55} & \ding{55} & \emph{L} & \ding{55} & \ding{51} & \ding{55} & \ding{55} & \ding{55} & 90 & DE\\
REVERIE~\cite{qi2020reverie}  & 22k & \ding{55} & \ding{55} & \ding{51} & \ding{55} & \ding{55} & \ding{55} & \emph{L} & \ding{55} & \ding{51} & \ding{55} & \ding{55} & \ding{55} & 90 & DE\\
OVON~\cite{yokoyama2024hm3d}   & 53k & \ding{51} & \ding{55} & \ding{55} & \ding{55} & \ding{55} & \ding{55} & \emph{L} & \ding{55} & \ding{55} & \ding{51} & \ding{55} & \ding{55} & 181 & CE\\
GOAT-Bench~\cite{khanna2024goat}  & 725k & \ding{51} & \ding{55} & \ding{55} & \ding{55} & \ding{51} & \ding{55} & \emph{V,L} & \ding{55} & \ding{55} & \ding{51} & \ding{55} & \ding{55} & 181 & CE\\
IR2R-CE~\cite{krantz2023iterative}   & 414 & \ding{55} & \ding{55} & \ding{51} & \ding{55} & \ding{55} & \ding{55} & \emph{L} & \ding{55} & \ding{51} & \ding{55} & \ding{55} & \ding{55} & 71 & CE\\
LHPR-VLN~\cite{song2024towards}  & 3.3k & \ding{51} & \ding{55} & \ding{51} & \ding{55} & \ding{55} & \ding{51} & \emph{L} & \ding{55} & \ding{55} & \ding{51} & \ding{55} & \ding{55} & 216 & CE\\
\hline
OctoNav-Bench & 45k & \ding{51} & \ding{51} & \ding{51} & \ding{51} & \ding{51} & \ding{51} & \emph{V,L,P} & \ding{51} & \ding{51} & \ding{51} & \ding{51} & \ding{51} & 438 & CE  \\
\hline
\end{tabular}
}

}
\vspace{-4mm}
\label{tab:benchmark}
\end{table}

\subsection{Data Analysis.}
\vspace{-0.5mm}
We illustrate comparisons between OctoNav-Bench and previous benchmarks as shown in Tab~\ref{tab:benchmark}. 
Generally, OctoNav-Bench pioneeringly unifies different tasks and contains free-form instruction, \ie, each instruction includes multi-modal (V,L,P) and multi-capability (noted as mixed). 
GOAT-Bench only includes ObjNav and Ins-ImgNav (goal-level instructions), lacking VLN capability that follows step-by-step action-level instructions. Moreover, each instruction in GOAT-Bench only involves a single modality and capability (non-mixed), making it a multiple set of independent tasks. LHPR-VLN includes only ObjNav and VLN, only involving the language modality and lacking the important visual-goal capability. Also, OctoNav-Bench is the pioneer work that provides TBA annotations.

\begin{figure}[t]
    \begin{center}
    \centerline{\includegraphics[width=1\linewidth]{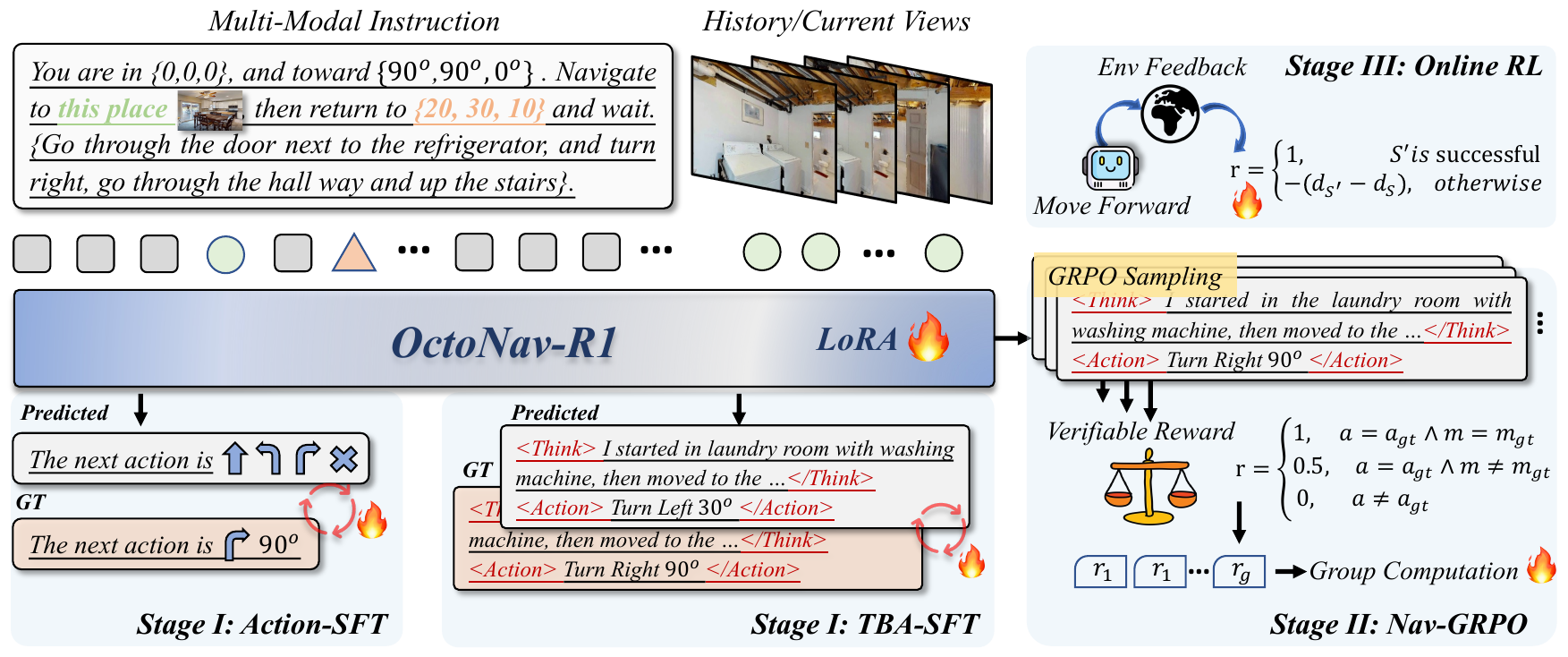}}
        \caption{\textbf{Overview of the HTP for training \emph{OctoNav-R1}.} The model takes multi-model instruction and visual observation as inputs, and produces textual answers, where model architecture details are in appendix~\ref{append:architecture}. HTP contains three training stages, which are described in Sec.~\ref{sec:method}.}
    \vspace{-6mm}
        \label{fig:octonav-r1}
    \end{center}
\end{figure}

\vspace{-1mm}
\section{OctoNav-R1}
\label{sec:method}
\vspace{-1mm}
We propose \emph{OctoNav-R1} to fully unleash the advantage offered by \emph{OctoNav-Bench}.
Overall, OctoNav-R1 is built upon LLaMA-VID~\cite{li2024llama} with specific architecture designs (appendix \ref{append:architecture}) to enable it to receive multi-modal instructions and produce low-level actions in an end-to-end manner.
Moreover, as illustrated in Fig.~\ref{fig:octonav-r1}, we propose a Hybrid Training Paradigm (HTP), which is a multi-stage reinforcement fine-tuning. 
Specifically, in \emph{Stage I}, we conduct imitation learning by leveraging the annotated data in OctoNav-Bench to apply supervised fine-tuning (SFT), which contains two phases.
First, in Action-SFT phase, the model is trained with instruction-trajectory pairs, enabling it to follow instructions.
Second, in TBA-SFT phase, the model is trained via TBA-CoT data, equipping the model with the ability to think before action.
In \emph{Stage II}, we propose Nav-GRPO with customized reward functions to further enhance the model's thinking capacity. 
In \emph{Stage III}, we conduct online RL in simulations supported by OctoNav-Bench, which enables trial-and-error and active learning.

\subsection{Action and TBA Supervised Fine-Tuning}

\vspace{-0.5mm}
\noindent\textbf{Action-SFT.}
\label{sec:sft}
In this phase, we train OctoNav-R1 (denoted as $\pi_\theta$) with instruction-trajectory pairs, enabling it to receive multi-modal inputs and produce navigation actions, as shown in Fig.~\ref{fig:octonav-r1}.
Technically, an instruction-trajectory pair contains multiple action steps within the trajectory. For each action step, a training sample can be presented as $(\mathcal V,\mathcal I, \mathcal A)$, where $\mathcal V$ is the visual observation, $\mathcal I$ is the instruction, and $\mathcal A$ refers to the answer. The visual observation $\mathcal V=(\mathcal V_{h},\mathcal V_{c})$ contains two parts, \ie, the historical observations (video) $\mathcal V_h\in \mathbb{R}^{N_h\times H\times W\times 3}$ and the current observation (image) $\mathcal V_c\in \mathbb{R}^{H\times W\times 3}$. $N_h$ is the frame number.
Besides, the instruction $\mathcal I$ contains multiple modalities, \ie, visual, textual, coordinate. We leverage placeholders to replace non-textual elements in the instruction. 
For instance, we utilize \emph{<ImageNav>} to replace the reference image in the instruction. Then, corresponding images are transferred into image embeddings by the visual encoder, which will replace the corresponding placeholders' embeddings (details in appendix~\ref{append:architecture}). 
Besides, as shown in Fig.~\ref{fig:octonav-r1}, the ground-truth answer $\mathcal A$ consists of an action $a$ and its magnitude $m$. Note that $a$ is chosen from \emph{\{move forward, turn left, turn right, stop\}}. $m$ indicates the distance (\eg, $25cm$) or angle (\eg, $90^\circ$) of the action.
Therefore, the training loss is computed as follows:
\begin{equation}
\small
\mathcal L_{act}(\theta)=-\mathbb E_{(\mathcal V,\mathcal I, \mathcal A)\sim D_{act}^{}}\frac{1}{|\mathcal A|}\sum_{t=1}^{|\mathcal A|}\log\pi_\theta(\mathcal A^t|\mathcal V, \mathcal I, \mathcal T_{act},\mathcal A^{<t}),
\end{equation}
where $\mathcal T_{act}$ is the prompt and $D_{act}$ is the dataset. 
$\mathcal A^{t}$ represents the $t$-th token within $\mathcal A$, and $\mathcal A^{<t}$ represents corresponding tokens preceding $\mathcal A^{t}$.

\noindent\textbf{TBA-SFT.}
Previous VLA-based methods for embodied navigation typically operate as black-box models, \ie, directly mapping multi-modal inputs to low-level actions without any explicit reasoning process. 
However, generalist-oriented navigation agents encounter complex and diverse tasks involving multi-modal and multi-capability instructions, thereby requiring reasoning ability.
Inspired by the think-before-answer within DeepSeek-R1~\cite{deepseekai2025deepseekr1incentivizingreasoningcapability}, we aim to bring a similar Think-Before-Action (TBA) paradigm to promote deliberative decision-making. 
Specifically, in this stage, we leverage the annotated TBA-CoT dataset in OctoNav-Bench to fine-tune the model. We aim to encourage the model to output in a structured format consisting of two clearly delineated segments, \ie, \emph{<Think> reasoning thoughts</Think><Action>executable actions</Action>}. As shown in Fig.~\ref{fig:octonav-r1}, a training sample is $(\mathcal V,\mathcal I, \mathcal A)$, where the answer $\mathcal A$ lies in the TBA format.
The training loss for this stage is:
\begin{equation}
\small
\mathcal L_{tba}(\theta)=-\mathbb E_{(\mathcal V,\mathcal I, \mathcal A)\sim D_{tba}^{}}\frac{1}{|\mathcal A|}\sum_{t=1}^{|\mathcal A|}\log\pi_\theta(\mathcal A^t|\mathcal V, \mathcal I, \mathcal T_{tba},\mathcal A^{<t}),
\end{equation}
where $\mathcal T_{tba}$ is the prompt and $D_{tba}$ is the TBA-CoT dataset.
Here, we use special prompts to instruct model to output in TBA format. Therefore, we can control the model to output direct action or in TBA format via different prompts. Thus, we can flexibly regulate the TBA frequency during inference.

\subsection{Nav-GRPO}
After the TBA-SFT, which is a cold-start training phase for enabling TBA ability, the model acquires an initial ability to generate outputs in the structured \emph{<Think>...</Think><Action>...</Action>} format. 
In this stage, we propose Nav-GRPO, \ie, group relative policy optimization (GRPO) for navigation, with customized reward functions to further enhance the model's thinking ability. 
First, we select $N_{G}$ training samples from OctoNav-Bench, forming $D_{GRPO}=\{(\mathcal V_i, \mathcal I_i)\}_{i=1}^{N_{G}}$. For each sample $(\mathcal V_i, \mathcal I_i)$, we utilize model trained after TBA-SFT to generate $G$ outputs: 
\begin{equation}
\{o_{i,j}\}_{j=1}^G \sim \pi_{\theta_{old}}(\mathcal V_i, \mathcal I_i,\mathcal T_g),
\end{equation}
where $\mathcal T_g$ is the prompt. 
As introduced in Sec.~\ref{sec:sft}, an output $o_{i,j}$ consists of an action $a_{i,j}$ and a magnitude $m_{i,j}$. Let $a_{gt}$ and $m_{gt}$ be the corresponding ground-truth action and magnitude, then the reward for $o_{i,j}$ is customized as:
\begin{equation}
\small
    r_{i,j}=\begin{cases}
    1,&a_{i,j}=a_{gt} \land m_{i,j}=m_{gt},\\
    0.5,&a_{i,j}=a_{gt} \land m\not =m_{gt},\\
    0,&a_{i,j}\not=a_{gt}.
    \end{cases}
    \label{reward:grpo}
\end{equation}
By setting up such a stepped reward function, the model is encouraged to learn from a broader range of reward signals, even when the generated answer is not completely precise. Then the advantage for $o_{i,j}$ is obtained via:
\begin{equation}
\small
\delta_{i,j}=\frac{r_{i,j}-\text{mean}(r_{i,1},r_{i,2},\cdots,r_{i,G})}{\text{std}(r_{i,1},r_{i,2},\cdots,r_{i,G})}.
\end{equation}
Following~\cite{deepseekai2025deepseekr1incentivizingreasoningcapability}, the loss for Nav-GRPO is computed as:
\begin{equation}
\small
\begin{aligned}
\mathcal{L}_{GRPO}(\theta)&=-\mathbb{E}{[(\mathcal V_i, \mathcal I_i)\sim D_{GRPO},\{o_{i,j}\}_{j=1}^G \sim \pi_{\theta_{old}}(\mathcal V_i, \mathcal I_i,\mathcal T_g)]}\\&\frac{1}{G}\sum_{j=1}^G\frac{1}{|o_{i,j}|}\sum_{t=1}^{|o_{i,j}|}[\min(c_1\cdot\delta_{i,j}, c_{2}\cdot\delta_{i,j})-\beta KL(\pi_{\theta}||\pi_{\theta_{SFT}})],
\end{aligned}
\end{equation}
where $\pi_{\theta_{SFT}}$ is the reference model to stabilize the training process, and $c_1,c_2$ are defined via:
\begin{equation}
\small
    c_1=\frac{\pi_\theta(o_{i,j}^{t}|\mathcal V_i, \mathcal I_i,\mathcal T_{g},o_{i,j}^{<t})}{\pi_{\theta_{old}}(o_{i,j}^{t}|\mathcal V_i, \mathcal I_i,\mathcal T_{g},o_{i,j}^{<t})}, c_2=\text{clip}(c_1,1-\varepsilon,1+\varepsilon).
\end{equation}

\subsection{Online Reinforcement Learning}
Previous VLA-based agents are predominantly trained via imitation learning.
In this stage, we take further steps by combining VLA-based models with online RL.
This is made feasible by OctoNav-Bench, which provides diverse and continuous environments within a unified simulation platform.

Technically, we apply the advantage actor-critic algorithm (A2C) as the learning policy. 
We utilize a linear network with a pooling layer as the critic model $V_{critic}$. It takes the hidden states of the last Transformer layer in the OctoNav-R1 as input and produces a score for corresponding states. 
Moreover, the reward function is designed considering both the distance change and the goal. 
Specifically, when the agent reaches the current goal successfully, reward $r_{on}=1$. 
Otherwise, the distance change to the current goal will be utilized to compute the reward. 
Formally, at state $\mathcal S$, the agent acts via $\mathcal A$ to reach the next state $\mathcal S'$, and the reward for this step is defined as follows:
\begin{equation}
\small
    r_{on}(\mathcal S, \mathcal A, \mathcal S')=\begin{cases}
    1, & \mathcal S'\text{ is successful},\\
    -(d_{\mathcal S'}-d_{\mathcal S}), & \text{otherwise},\\
    \end{cases}
\end{equation}
where $d_{\mathcal S},d_{\mathcal S'}$ are the distances to the current goal at state $\mathcal S$ and $\mathcal S'$ respectively. For non-moving actions (\eg, turn right), the distance to the goal remains unchanged after execution. Thus, we make the agent slightly move forward $d'cm$ after each non-moving action, to obtain non-zero effective rewards. Formally, we adopt the temporal difference algorithm to train OctoNav-R1, and the loss function is achieved via:
\begin{equation}
\small
\mathcal L_{on}(\theta)=-\mathbb E_{(\mathcal S,\mathcal A, \mathcal S')\sim\pi_\theta}\frac{1}{|\mathcal A|}\sum_{t=1}^{|\mathcal A|}[r_{on}(\mathcal S,\mathcal A, \mathcal S')+\gamma V_{critic}(\mathcal S')- V_{critic}(\mathcal S)]\log\pi_\theta(\mathcal A^t|\mathcal S,\mathcal A^{<t}) ,
\end{equation}
where $\gamma$ is the discount factor. Besides, the critic model $V_{critic}$ is updated by the MSE loss:
\begin{equation}
\small
\mathcal L_{critic}(V_{critic})=\mathbb E_{(\mathcal S,\mathcal A, \mathcal S')\sim\pi_\theta}[r_{on}(\mathcal S,\mathcal A, \mathcal S')+\gamma V_{critic}(\mathcal S')- V_{critic}(\mathcal S)]^2.
\end{equation}
In practice, we adopt a warm-up training strategy, \ie, only train the critic model and freeze the parameters of OctoNav-R1 during the initial training phase.

\section{Experiments}
\label{sec:exp}
\vspace{-1mm}
\subsection{Experimental Setting}
\vspace{-1mm}
\begin{table*}[!t]
\centering
\caption{\textbf{Comparison with previous methods on test set.}
* denotes model modification and fine-tuning on OctoNav-Bench for comparison. \textdagger indicates fine-tuning on OctoNav-Bench.}
\vspace{-1mm}
\resizebox{\textwidth}{!}{
{
\renewcommand{\arraystretch}{1.35}
\setlength{\tabcolsep}{5pt}
\begin{tabular}{r||ccc||ccc||ccc||ccc||ccc||ccc}
\hline \thickhline

 \rowcolor[HTML]{f8f9fa} \multicolumn{1}{c||}{}& \multicolumn{3}{c||}{Overall} & \multicolumn{3}{c||}{Ins-ImgNav} & \multicolumn{3}{c||}{ImgNav} & \multicolumn{3}{c||}{PointNav} & \multicolumn{3}{c||}{ObjNav} & \multicolumn{3}{c}{VLN}\\
\rowcolor[HTML]{f8f9fa} \multicolumn{1}{c||}{\multirow{-2}{*}{Methods}} & SR& SPL & OSR & SR& SPL & OSR & SR& SPL & OSR & SR& SPL & OSR & SR& SPL & OSR & SR& SPL & OSR\\
\hline
\hline
\rowcolor[HTML]{f8f9fa} \multicolumn{19}{l}{\emph{MLLMs as Agent}} \\
Qwen-VL\cite{bai2025qwen2} & 0.00 & 0.00 & 2.00 & 0.00 & 0.00 & 0.00 & 0.00 & 0.00 & 0.00 & 0.00 & 0.00 & 0.00 & 0.41 & 0.16 & 6.15 & 2.86 & 2.86 & 2.86\\
Video-LLaVA~\cite{lin2024video} & 0.80 & 0.45 & 3.80 & 2.82 & 1.78 & 3.63 & 0.41 & 0.17 & 0.41 & 1.20 & 0.63 & 1.20 & 3.28 & 2.67 & 12.30 & 5.71 & 4.51 & 5.71\\
LLaVA-NeXT~\cite{zhang2024llavanextvideo} & 0.20 & 0.18 & 2.40 & 2.02 & 1.46 & 2.42 & 0.00 & 0.00 & 0.41 & 0.00 & 0.00 & 0.00 & 2.46 & 2.11 & 9.43 & 2.86 & 2.86 & 2.86\\
\hline 
\rowcolor[HTML]{f8f9fa} \multicolumn{19}{l}{\emph{Methods for DE}} \\
NaviLLM*~\cite{zheng2024towards} & 1.20 & 1.20 & 4.20 & 3.63 & 3.57 & 4.03 & 0.00 & 0.00 & 0.41 & 0.80 & 0.75 & 0.80 & 4.10 & 3.49 & 13.11 & 8.57 & 8.57 & 14.29\\
NavGPT-2*~\cite{zhou2024navgpt2} & 2.00 & 1.35 & 5.20 & 2.82 & 2.21 & 3.23 & 1.65 & 0.89 & 2.07 & 2.79 & 1.81 & 2.79 & 6.15 & 4.85 & 15.57 & 11.43 & 11.43 & 17.14\\
\hline 
\rowcolor[HTML]{f8f9fa} \multicolumn{19}{l}{\emph{Methods for CE}} \\
NaVid~\cite{zhang2024navid} & 5.80 & 4.34 & 11.40 & 10.48 & 7.35 & 12.10 & 5.37 & 3.80 & 6.20 & 7.57 & 5.85 & 7.97 & 23.36 & 17.95 & 37.30 & 25.71 & 25.29 & 37.14\\
Uni-NaVid~\cite{zhang2025uninavid} & 8.60 & 5.79 & 17.60 & 15.32 & 10.25 & 16.53 & 11.16 & 7.41 & 12.81 & 9.56 & 5.86 & 11.55 & 36.48 & 25.48 & 56.97 & 28.57 & 22.33 & 42.86\\
NaVid\textdagger~\cite{zhang2024navid} & 8.80 & 7.20 & 13.80 & 16.94 & 13.23 & 18.15 & 10.74 & 9.33 & 11.57 & 9.96 & 9.52 & 11.55 & 27.05 & 23.21 & 38.52 & 20.00 & 19.57 & 25.71\\
Uni-NaVid\textdagger~\cite{zhang2025uninavid} & 9.20 & 6.21 & 17.80 & 19.35 & 13.44 & 22.58 & 10.33 & 6.79 & 11.98 & 11.55 & 7.40 & 13.15 & 42.62 & 28.70 & 61.07 & 25.71 & 21.45 & \textbf{45.71}\\
\hline
OctoNav-R1~(ours) & \textbf{19.40} & \textbf{13.77} & \textbf{29.40} & \textbf{30.24} & \textbf{20.77} & \textbf{35.48} & \textbf{23.97} & \textbf{17.49} & \textbf{27.27} & \textbf{23.51} & \textbf{14.35} & \textbf{27.89} & \textbf{49.18} & \textbf{37.79} & \textbf{67.21} & \textbf{37.14} & \textbf{33.56} & 42.86\\
\hline
\end{tabular}
}
}
\vspace{-3mm}
\label{comparison}
\end{table*}
\vspace{-0.5mm}
\noindent\textbf{Environments.}
OctoNav-Bench is built upon Habitat simulator~\cite{habitat19iccv}. Scenes are diversely collected from MP3D, HM3D, Gibson, and ProcTHOR, including $400+$ and $40+$ scenes for train and test splits. Scenes used in test are unseen in train split. Annotated instruction-trajectory pairs are $45k+$, and TBA-CoT contains $10k+$ instruction-think-action pairs. 
Real world deployment in appendix~\ref{append-real}.

\vspace{-0.5mm}
\noindent\textbf{Metrics.} We utilize success rate (SR), oracle success rate (OSR), and the success weighted by path length (SPL). 
The success of a task means all sub-tasks within are successful in order, while the oracle success only concerns whether the final goal is reached.
The success of a sub-task means the current and predecessor sub-tasks are completed in order.
Note that each sub-task corresponds to a specific capability, thus we can calculate the metrics for each capability. More details in appendix~\ref{append:bench-metric}.

\vspace{-1mm}
\subsection{Comparison with Previous Methods}
\vspace{-1mm}
Task-specific models designed for the single navigation task/capability can not achieve our setting, as each instruction in OctoNav-Bench involves multi-modal and multi-capability. Therefore, we adopt LLM-/MLLM-based models~\cite{zheng2024towards,zhou2024navgpt2,zhang2024navid,zhang2025uninavid} for comparison, since they possess a certain degree of generalized ability (shown in Tab.~\ref{comparison}).
First, we prompt off-the-shelf MLLMs (\eg, Qwen-VL) as agents to choose actions. 
While video-based MLLMs perform better than image-based ones, \eg, Video-LLaVA achieve $0.80\%$ overall SR, the performance is still poor for such complex tasks.

Original NaviLLM and NavGPT-2 can perform VLN, but only for discrete environments (DE).
We carefully revise them (marked * in Tab~\ref{tab:benchmark}) \eg, modifying the waypoint-selection head to action-selection head, to enable them to move in CE. 
Then we also fine-tune them on OctoNav-Bench. 
The performance is better than off-the-shelf MLLMs, yet still low, \eg, NavGPT-2* achieves $2.00\%$ SR. It indicates the huge gap between task settings and between the discrete and continuous environments.

NaVid and Uni-NaVid are originally trained via collected multi-task datasets.
However, the low performance confirms the essential difference between OctoNav-Bench and multi-task dataset that are simply collected from existing datasets. 
Further, we fine-tune and evaluate NaVid\textdagger and Uni-NaVid\textdagger. The best performance $9.20\%$ SR obtained via Uni-NaVid\textdagger is still far lower than our OctoNav-R1 model. It confirms our technical contribution of model design (appendix~\ref{append:architecture}) and HTP (Sec.~\ref{sec:method}).

\subsection{Ablation Study and Visualizations}
We investigate the contribution of each proposed component via extensive ablations. 
The results are shown in Tab.~\ref{tab:ablation-main} and Tab.~\ref{tab:ablation-mini}. Note that the Base-model's architecture in appendix~\ref{append:architecture}.
Base-model achieves higher performance on  ObjNav and VLN, while other capabilities are not satisfied,
\eg, $5.37\%$ SR on ImgNav and $7.57\%$ SR on PointNav.

\begin{table*}[t]
\centering
\caption{\textbf{Main Ablations of the \emph{HTP}.} The performance is gradually improved with the continuous addition of the proposed methods in each stage.}
\resizebox{\textwidth}{!}{
{
\renewcommand{\arraystretch}{1.3}
\setlength{\tabcolsep}{5pt}
\begin{tabular}{r||ccc||ccc||ccc||ccc||ccc||ccc}
\hline \thickhline

 \rowcolor[HTML]{f8f9fa} \multicolumn{1}{c||}{}& \multicolumn{3}{c||}{Overall} & \multicolumn{3}{c||}{Ins-ImgNav} & \multicolumn{3}{c||}{ImgNav} & \multicolumn{3}{c||}{PointNav} & \multicolumn{3}{c||}{ObjNav} & \multicolumn{3}{c}{VLN}\\
\rowcolor[HTML]{f8f9fa} \multicolumn{1}{c||}{\multirow{-2}{*}{Methods}} & SR& SPL & OSR & SR& SPL & OSR & SR& SPL & OSR & SR& SPL & OSR & SR& SPL & OSR & SR& SPL & OSR\\
\hline
\hline
Base-Model & 5.80 & 4.34 & 11.40 & 10.48 & 7.35 & 12.10 & 5.37 & 3.80 & 6.20 & 7.57 & 5.85 & 7.97 & 23.36 & 17.95 & 37.30 & 25.71 & 25.29 & 37.14\\
\hline
+Action-SFT & 8.80 & 7.20 & 13.80 & 16.94 & 13.23 & 18.15 & 10.74 & 9.33 & 11.57 & 9.96 & 9.52 & 11.55 & 27.05 & 23.21 & 38.52 & 20.00 & 19.57 & 25.71\\ 
+TBA-SFT & 14.40 & 10.32 & 25.00 & 25.40 & 16.57 & 31.05 & 19.01 & 14.56 & 23.14 & 21.51 & 14.93 & 25.10 & 42.21 & 30.67 & 61.07 & 31.43 & 26.46 & \textbf{48.57}\\
+Nav-GRPO & 17.00 & 12.04 & 25.00 & 29.44 & 20.73 & 33.47 & 19.42 & 14.27 & 22.73 & 21.51 & \textbf{16.37} & 24.70 & 48.77 & 35.63 & 62.70 & 34.29 & 29.14 & \textbf{48.57}\\
+Online RL & \textbf{19.40} & \textbf{13.77} & \textbf{29.40} & \textbf{30.24} & \textbf{20.77} & \textbf{35.48} & \textbf{23.97} & \textbf{17.49} & \textbf{27.27} & \textbf{23.51} & 14.35 & \textbf{27.89} & \textbf{49.18} & \textbf{37.79} & \textbf{67.21} & \textbf{37.14} & \textbf{33.56} & 42.86\\
\hline
\end{tabular}
}

}
\label{tab:ablation-main}
\end{table*}

\begin{table*}[t]
\centering
\caption{\textbf{Ablations.} We conduct detailed ablations in terms of different aspects.}
\vspace{-1mm}
\resizebox{\textwidth}{!}{
\subfloat[Reward design in Nav-GRPO. \label{tab:ablation:reward}]{
   \renewcommand{\arraystretch}{1.1}
    \setlength{\tabcolsep}{5pt}
   \begin{tabular}{r||ccc}
\hline \thickhline

 \rowcolor[HTML]{f8f9fa} \multicolumn{1}{c||}{Reward}& \multicolumn{3}{c}{Overall} \\
\rowcolor[HTML]{f8f9fa} \multicolumn{1}{c||}{Type} & SR& SPL & OSR \\
\hline
\hline
Strict & 16.20 & 11.74 & 25.40 \\
Loose & 15.40 & 10.97 & \textbf{26.00} \\ 
Stepped & \textbf{17.00} & \textbf{12.04} & 25.00 \\
\hline
\end{tabular}\hspace{6mm}
}
\subfloat[Prompt template ablations in Nav-GRPO.  \label{tab:ablation:grpo}]{
      \renewcommand{\arraystretch}{1.1}
    \setlength{\tabcolsep}{5pt}
   \begin{tabular}{r||ccc}
\hline \thickhline

 \rowcolor[HTML]{f8f9fa} \multicolumn{1}{c||}{}& \multicolumn{3}{c}{Overall} \\
\rowcolor[HTML]{f8f9fa} \multicolumn{1}{c||}{\multirow{-2}{*}{Template Type}} & SR& SPL & OSR \\
\hline
\hline
Single Template & 15.80 & 10.93 & \textbf{25.40} \\
Diverse Templates & \textbf{17.00} & \textbf{12.04} & 25.00 \\ 
\hline
\end{tabular}\hspace{6mm}
}
\subfloat[Thinking frequency ablations.  \label{tab:ablation:think}]{
      \renewcommand{\arraystretch}{1.1}
    \setlength{\tabcolsep}{5pt}
   \begin{tabular}{r||ccc}
\hline \thickhline

 \rowcolor[HTML]{f8f9fa} \multicolumn{1}{c||}{Thinking}& \multicolumn{3}{c}{Overall} \\
\rowcolor[HTML]{f8f9fa} \multicolumn{1}{c||}{Frequency} & SR& SPL & OSR \\
\hline
\hline
per 10 steps& 17.00 & 11.51 & 27.60 \\
per 20 steps& \textbf{19.40} & \textbf{13.77} & \textbf{29.40} \\ 
per 40 steps& 18.80 & 12.62 & 28.20 \\
\hline
\end{tabular}
}

}
\vspace{-2mm}
\label{tab:ablation-mini}
\end{table*}

\noindent\textbf{Action-SFT.} 
As shown in Tab.~\ref{tab:ablation-main}, after the Action-SFT with instruction-trajectory pairs of OctoNav-Bench, the model improves shortage capabilities and increases the overall SR (from $5.80\%$ to $8.80\%$). 
However, the performance on VLN capability is slightly dropped, \eg, SR from $25.71\%$ to $20.00\%$, indicating the VLN capability is sacrificed to improve the generalist ability.

\begin{figure}[H]
    \begin{center}
    \vspace{-3mm}
    \centerline{\includegraphics[width=0.8\linewidth]{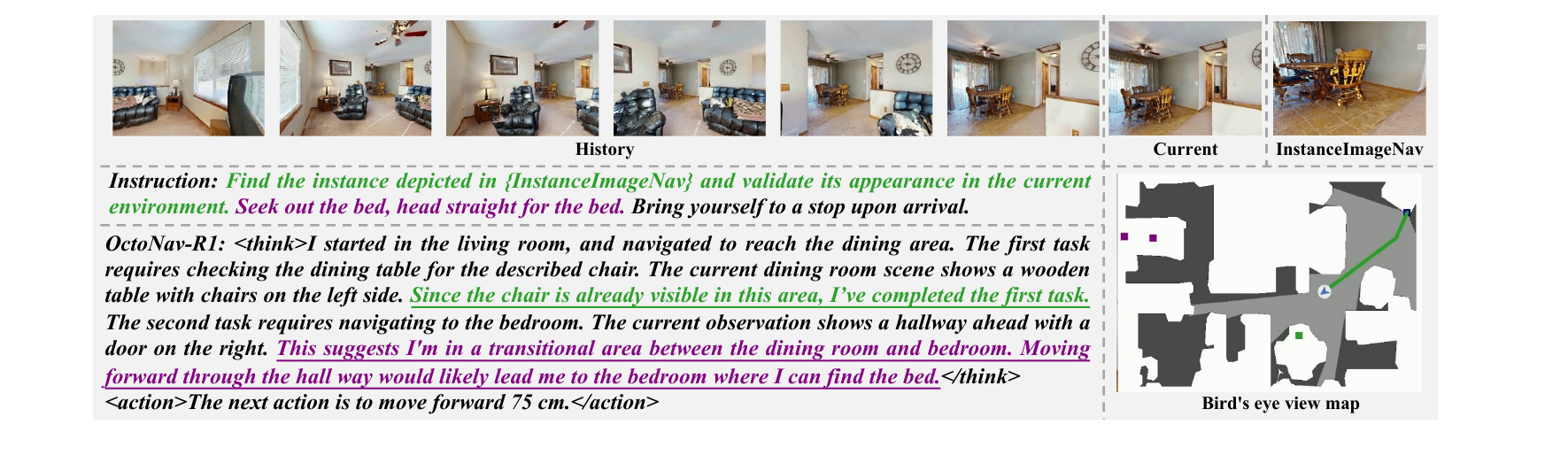}}
    \vspace{-2mm}
        \caption{{Visualization of TBA in a trajectory.}}
    \vspace{-5mm}
        \label{fig:think-vis}
    \end{center}
\end{figure}
\vspace{-0.5mm}
\vspace{-0.5mm}
\noindent\textbf{TBA-SFT.} 
After the TBA-SFT phase, the overall SR is improved by $5.60\%$, demonstrating the effectiveness of the thinking process. Moreover, the model can better handle multi-capability with the thinking ability, as the performance is greatly improved in terms of all capabilities. 
For example, the SR is improved by $11.55\%$ on PointNav and by $15.16\%$ on ObjNav.

\vspace{-0.5mm}
\noindent\textbf{Nav-GRPO.} As shown in Tab.~\ref{tab:ablation-main}, with Nav-GRPO, the overall SR is improved from $14.40\%$ to $17.00\%$. Besides, the SPL of VLN is improved to $29.14\%$, indicating Nav-GRPO can further boost the thinking quality, leading to a better ability for complex VLN tasks. 
We further investigate the influence of reward design and prompt diversity. In Tab.~\ref{tab:ablation:reward}, `Strict' means the reward of the partially correct answer ($a=a_{gt}\land m\not= m_{gt}$) is $0$, and `Loose' means the reward is $1$. As shown, the stepped version achieves the best performance.
Besides, as shown in Tab.~\ref{tab:ablation:grpo}, diverse prompts can enhance the generalization ability of the model, and thereby improving SR.

\vspace{-0.5mm}
\noindent\textbf{Online RL.} 
As shown in Tab.~\ref{tab:ablation-main}, the performance is further enhanced via online RL, where SR is improved by $2.85\%$ on VLN and by $4.55\%$ on ImgNav. 
Besides, the overall SPL is improved by $1.73\%$, showing that the model learns a more efficient navigation strategy.

\vspace{-0.5mm}
\noindent\textbf{Thinking Frequency.} 
We investigate the thinking frequency in Tab.~\ref{tab:ablation:think}.
Overall, the performance fluctuations are not sensitive, \eg, $19.40\%$ SR of per-$20$-step \emph{v.s} $18.80\%$ SR of per-$40$-step. However, where and when to think is still a valuable topic, and we leave it for future work.

\noindent\textbf{Visualization of TBA.}
As shown in Fig.~\ref{fig:think-vis}, since chair is already in the current view, the agent realizes the first task is completed, and continues to plan for next tasks. 
Such thinking shows that OctoNav-R1 can understand task order, evaluate, and switch states. 
Besides, the agent can infer the location of the bed based on the common sense of indoor layout. 
More visualizations are in the appendix.
\vspace{-1mm}
\section{Conclusion}
\label{sec:conclusion}
\vspace{-1mm}
In this work, we introduce OctoNav-Bench and OctoNav-R1, aiming to build generalist agents capable of following free-form instructions with multi-modal and multi-capability. 
OctoNav-Bench is constructed with an annotation pipeline, providing instruction-trajectory pairs and a TBA-CoT dataset to capture reasoning processes. 
For OctoNav-R1, we design a VLA-type model that generates low-level actions purely from 2D visual observations. 
We propose HTP with three stages: Action-/TBA-SFT, Nav-GPRO, and Online RL. Our method emphasizes thinking-before-action, showing reasoning and generalization. Experiments confirm the superiority of OctoNav-R1 over prior methods.

\medskip

{
\small
\bibliographystyle{unsrt}
\bibliography{reference}
}

\newpage


\appendix

\section{Appendix: More Details of OctoNav-Bench}
\label{append:bench}
\vspace{-1mm}
\subsection{More Related Works}

\vspace{-0.5mm}
\noindent\textbf{Embodied Navigation Tasks and Benchmarks.}
Embodied navigation aims to make agents perceive and move toward target locations in unseen environments.
Existing research primarily centers around specific tasks
including ObjNav~\cite{chaplot2020object,gao2023room,gao2021room}, PointNav~\cite{habitat2020sim2real}, ImgNav~\cite{sun2023fgprompt}, Ins-ImgNav~\cite{habitatchallenge2023,krantz2023navigating,krantz2022instance}, and VLN~\cite{anderson2018vision,krantz2020beyond,qi2020reverie,liu2023bird,liu2024vision,liu2024volumetric,gao2023adaptive,chen2022reinforced,zhao2022target}.
which differ in both objective and instruction modality. 
For instance, PointNav requires reaching a specified coordinate with the instruction given as a coordinate. ImageNav involves locating an object or region depicted in a reference image, where the instruction lies in visual modality. 
Correspondingly, benchmarks are developed for the development and evaluation of each task, \eg, R2R-CE~\cite{krantz2020beyond} and REVERIE~\cite{qi2020reverie} for the VLN task. 
Although these task-specific benchmarks have significantly advanced progress, their strict segregation creates obstacles for developing generalist navigation agents.
Models trained with narrowly defined tasks struggle to adapt to diverse, open-ended instructions, inherently limiting their versatility and scope.
In this work, we pioneeringly introduce the OctoNav-Bench. It features free-form instructions, where each instruction may span multiple modalities and potentially specify composite navigation capabilities.
By unifying multiple modalities and capabilities and allowing flexible instruction formats, we aim to catalyze research toward generalist agents that can interpret arbitrary commands and execute a wide range of embodied navigation goals.
\subsection{More Construction Details}
\label{append:bench-construction}
\vspace{-0.5mm}
\subsubsection{Ins-ImgNav Goal Generation}
\begin{figure}[H]
    \begin{center}
    \centerline{\includegraphics[width=0.9\linewidth]{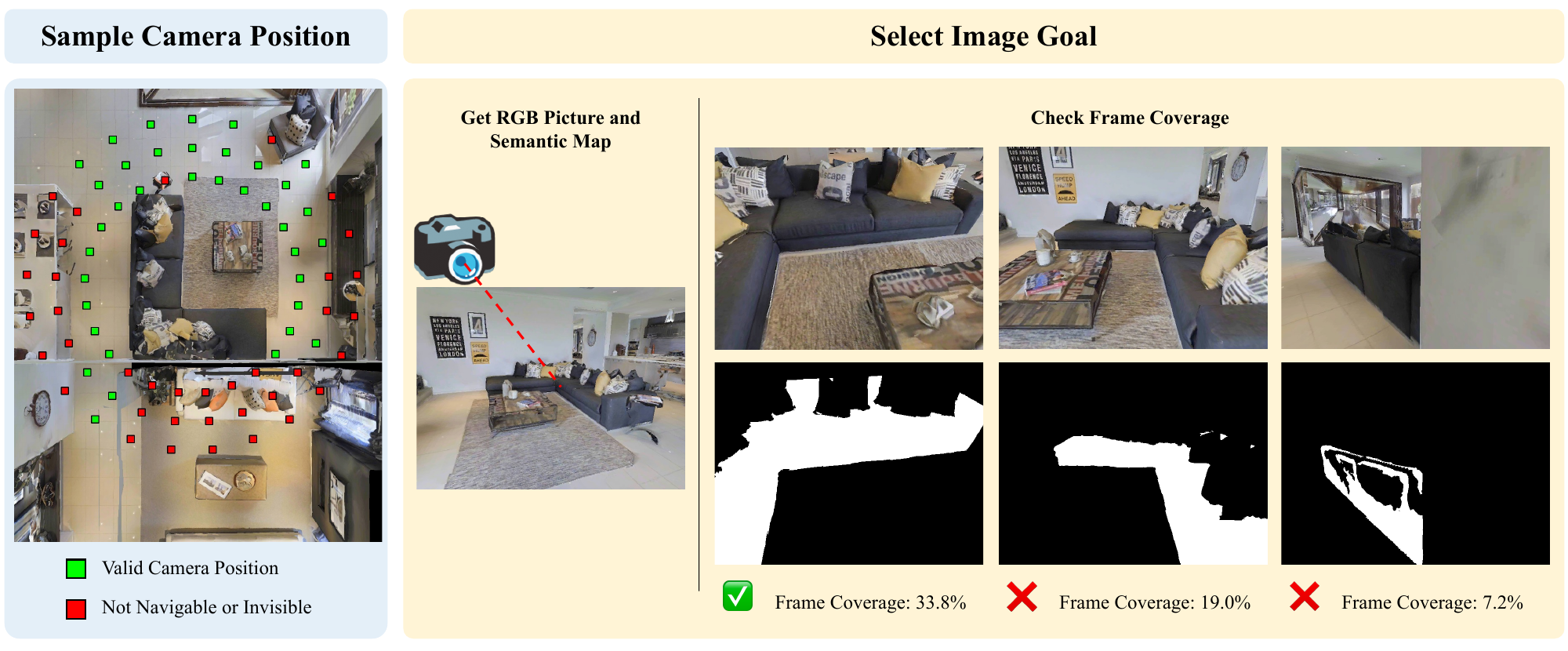}}
    \vspace{-1mm}
        \caption{Construction pipeline of instance image goals for Ins-ImgNav}
    \vspace{-3mm}
        \label{fig:ins-img}
    \end{center}
\end{figure}

We mainly follow the generation process in \cite{krantz2022instance}.
The generation of instance image goals proceeds in two stages, \ie, sample camera position and select image goal, as shown in Fig.~\ref{fig:ins-img}. 

\noindent\textbf{Sampling Camera Positions.}
During camera position sampling, $36$ candidate points are uniformly distributed along concentric circles centered at the bounding box centroid of the target object (\ie, evenly spaced points every $10$ degrees), where radius of circles $r \in {0.5m, 1.0m, 1.5m, 2.0m}$. For each candidate point, five height values are randomly sampled from the range of $[0.8,1.5]$ meters to form potential camera positions. All candidate camera positions undergo validity assessment based on two criteria: navigability and visibility. Positions occupied by other objects or walls are considered not navigable, while positions not able to see the target object are marked as invisible. Only navigable and visible positions (green rectangles in the figure) are selected and advance to the next stage.

\noindent\textbf{Selecting Image Goal.}
From the validated camera position set, we subsequently select positions that optimally frame the target object. To ensure visual prominence, each camera is oriented along the vector from the camera position to the object's center, thereby maximizing the object's projection in the image. We then compute the frame coverage for each position, defined as the ratio of image pixels occupied by the target object. This metric is calculated from the semantic map of the current scene. Positions whose frame coverage exceeds a threshold of $20\%$ are retained as Ins-ImgNav goal camera positions, and the RGB picture taken from these positions is served as Ins-ImgNav goal.

\subsubsection{Trajectory Generation}
\label{append:bench-construction-traj}
During trajectory generation, we apply several constraints to ensure diversity and reality of trajectories.
Firstly, the starting and ending points of trajectories are randomly sampled within the navigable areas of the scene, with inaccessible start-end pairs being eliminated. Subsequently, the Euclidean distance and geodesic distance of each trajectory are computed. We filter trajectories based on geodesic distance to ensure appropriate path distance. For capability combinations containing $k$ capabilities, the generated trajectory distances are constrained within the interval $[3k, 10k]$ meters.
Additionally, to encourage moderately complex trajectories and avoid straight paths, we define a straight ratio as the quotient of geodesic distance divided by Euclidean distance. A higher ratio indicates greater deviation from straight path. Trajectories with ratios exceeding $1.1$ are selected as candidate paths. For trajectories with ratios within $[1.0, 1.1]$, we employ a probabilistic selection criterion defined by $10\times(ratio - 1)^2$ to mitigate excessive linear trajectories while preserving stochasticity.

For VLN capability, we first include the VLN data of R2R-CE and RxR-CE, then generate some VLN data using C-INSTRUCTOR~\cite{kong2024controllable}. Since instructions and trajectories are strictly aligned, our generated trajectories must fully encompass the complete VLN path. Randomly generated trajectories, as previously described, rarely align perfectly with VLN trajectories, necessitating special trajectory generation for VLN.
To deal with VLN, we generate trajectory segments before and after the VLN path, then concatenate them into a complete trajectory. These trajectory segments are generated using the algorithm mentioned before, with either the starting or ending point fixed to match the VLN trajectory. During concatenation, trajectories must be screened to ensure natural continuity with the VLN path, avoiding moving back and forth or $180\circ$ turns.
Let $S$ denote the starting point and $T$ the endpoint of the VLN trajectory. For the former trajectory segment (starting at $A$ and ending at $S$) and the latter segment (starting at $T$ and ending at $B$), distance-based ratios are employed as filtering criteria. A trajectory is deemed sufficiently natural if it satisfies the following conditions: $(dis_{AS}+dis_{ST})/dis_{AT}\le 3$, $(dis_{ST}+dis_{TB})/dis_{SB}\le 3$, $(dis_{AS}+dis_{ST}+dis_{TB})/dis_{AB} \le 5$. Trajectories meeting all three criteria are added to the candidate pool. This approach ensures geometric coherence, minimizes unnatural motions, and maintains alignment with VLN trajectories.

\begin{figure}[t]
    \begin{center}
    \centerline{\includegraphics[width=0.8\linewidth]{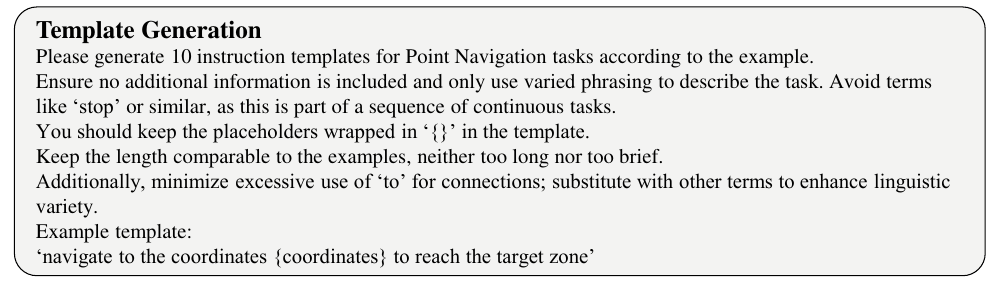}}
    \vspace{-1mm}
        \caption{Example Prompt for Template Generation.}
    \vspace{-3mm}
        \label{fig:prompt_gen_template}
    \end{center}
\end{figure}

\subsubsection{Instruction Template Generation}
\noindent\textbf{Template Pool Building.}
We leverage the GPT to generate several templates in advance, building a template pool. We wrote one example for each type of template, and ask GPT to rewrite it, the prompt is shown in Fig. ~\ref{fig:prompt_gen_template}. Each generated template is checked by human. The most important part is instruction of capabilities, we prepare $10$ templates for each capability. For example, `proceed toward the coordinates {coordinates} until reaching the designated area' for PointNav, `navigate to the object shown in {InstanceImageNav} and confirm its visibility in the present scene' for Ins-ImgNav. However, the VLN instruction is a complete passage generated with corresponding trajectory, so templates are not required. In addition, a $10$ series of conjunctions is generated to concatenate capability instructions and improve diversity. Moreover, the template pool includes $10$ various stop instruction, such as 'stop after reaching the target.'

\noindent\textbf{Instruction Template Forming.}
The instruction templates include three parts: the initial coordinate and orientation of agent, a sequence of task instructions and a stop command. The first part is fixed, showing the inital state of agent: `Your current position is (x, y, z) and your current orientation is (dx, dy, dz)'. The second part is the main body of instruction. We concatenate template of single capability instructions into a sequence, which are all randomly selected from template pool. Among the sequences, half of them are concatenated with conjunctions, while others are not. The last part indicates that the agent need to stop after finishing all the tasks, also randomly selected. Finally, we link the three part to form a complete instruction template. Through randomization, we generated a large number of diverse instructions. 

\subsubsection{PointNav Setting}
In the PointNav task of the Habitat Challenge 2019, agents were equipped with GPS and compass sensors, enabling real-time localization. However, Habitat Challenge 2020 discontinued this configuration, as high-precision sensors are generally unavailable in indoor environments. Following the Habitat Challenge setting, we do not provide real-time positional feedback. Instead, we provide navigation instructions including the starting coordinates and orientation and the target coordinates specified in the PointNav task.
We employ absolute coordinates of the scene's intrinsic coordinate system rather than relative coordinates. This design facilitates the model's ability to learn spatial information and construct a coherent spatial coordinate system. If relative coordinates were used, positional error is more likely to accumulate during trajectories requiring frequent turns, making it harder to find the goal position.

\subsubsection{Instruction Extension and Quality Check.}
\label{append:bench-construction-check}
\noindent\textbf{Instruction Extension.}
We aim to enhance not only the diversity of navigation capabilities but also the linguistic variability of instructions. 
Thus, we leverage DeepSeek to rewrite some instructions into three distinct variants.
All three versions convey the same core intent and task objective, differing only in their linguistic form. 
As a result, trajectories are paired with a triplet of semantically equivalent but stylistically diverse instructions, thereby enriching the dataset's linguistic coverage and promoting the model's generalization to diverse language expressions.

\noindent\textbf{Quality Check.}
To further improve data quality, we introduce a semi-automatic verification stage. 
First, we employ GPT to perform automatic filtering, targeting common issues such as semantic inconsistencies among the instruction variants, hallucinations, or vague task goals. 
Following this, humans manually review the filtered outputs to either correct minor or ambiguous errors or discard examples that are difficult to revise. 
This process ensures that only high-quality and well-grounded data are retained, enhancing the overall reliability and robustness of the dataset.

\subsubsection{Think-Before-Action Chain-of-Thoughts Generation Pipeline.}
Our TBA-CoT generation pipeline includes three stages: image information extraction, historical information aggregation, and TBA reasoning. The first two stages leverage multimodal models (\eg, Qwen-VL), while the final stage utilizes reasoning models (\eg, DeepSeek-R1).

\begin{figure}[H]
    \begin{center}
    \centerline{\includegraphics[width=0.8\linewidth]{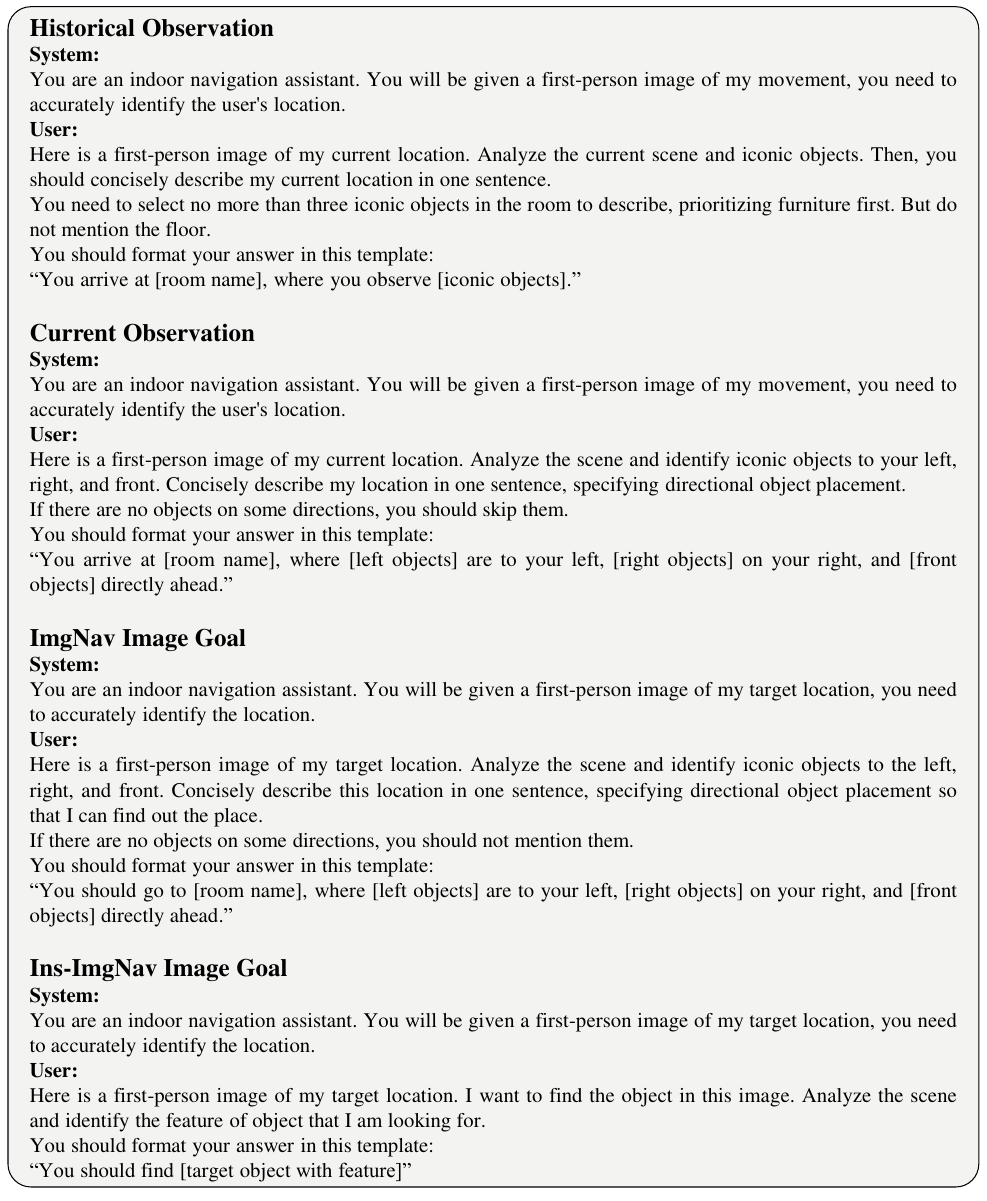}}
    \vspace{-1mm}
        \caption{Prompt for Image Information Extraction.}
    \vspace{-3mm}
        \label{fig:prompt_observation}
    \end{center}
\end{figure}

\noindent\textbf{Image Information Extraction.}
The image information extraction process involves four image categories: current observation images, historical observation image sequences, ImgNav goal images, and Ins-ImgNav goal images. Each category requires distinct extraction strategies through customized prompt engineering. For historical observation images, we primarily extract room types and their distinctive landmark objects. Current observation images and ImgNav goal images necessitate the extraction of both room types and spatial relationships (\eg, objects positioned to the left, right, or front) to decide next move or identify goal position. Ins-ImgNav processing focuses on object attributes such as color characteristics and adjacent object relationships to facilitate target identification. The detailed prompts are shown in Fig.~\ref{fig:prompt_observation}.

\noindent\textbf{Historical Information Aggregation.}

\begin{figure}[H]
    \begin{center}
    \centerline{\includegraphics[width=0.8\linewidth]{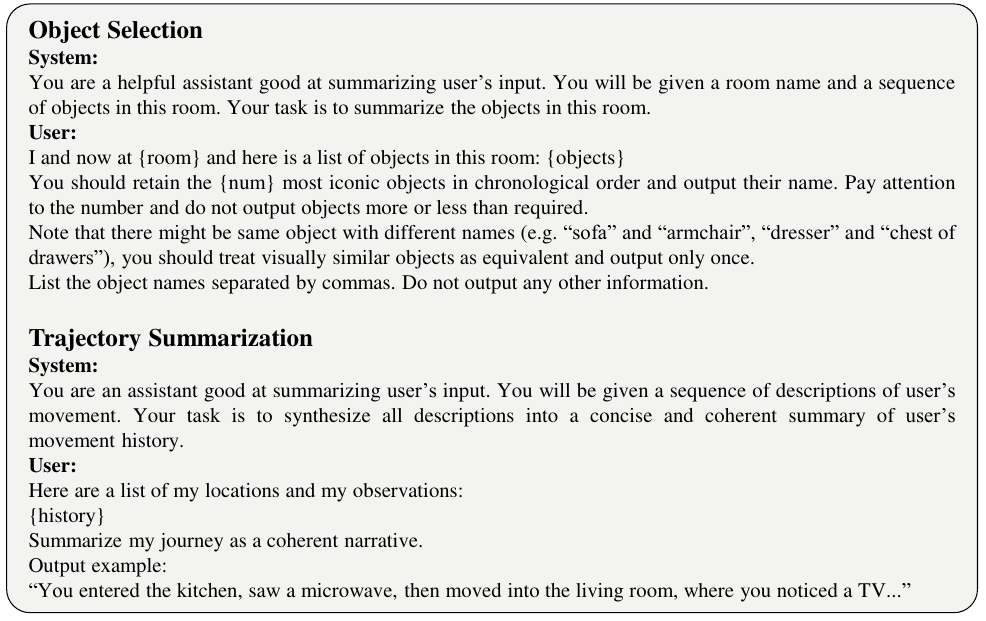}}
    \vspace{-1mm}
        \caption{Prompt for Historical Information Aggregation.}
    \vspace{-3mm}
        \label{fig:prompt_summerization}
    \end{center}
\end{figure}

In the historical information aggregation phase, we initially collect all objects identified from historical observations. To address information redundancy particularly in expansive areas (\eg, dining areas, hallways), we implement an object selection mechanism through the multimodal model, retaining the most iconic $30\%$ of the objects. At the same time, we prompt the model to merge synonyms to resolve naming inconsistencies (\eg, `sofa' vs. `armchair') that may occur during initial stage. The filtered object sets are then concatenated and input into the model to generate structured historical trajectory summaries. The detailed prompts are shown in Fig.~\ref{fig:prompt_summerization}.

\begin{figure}[H]
    \begin{center}
    \centerline{\includegraphics[width=0.85\linewidth]{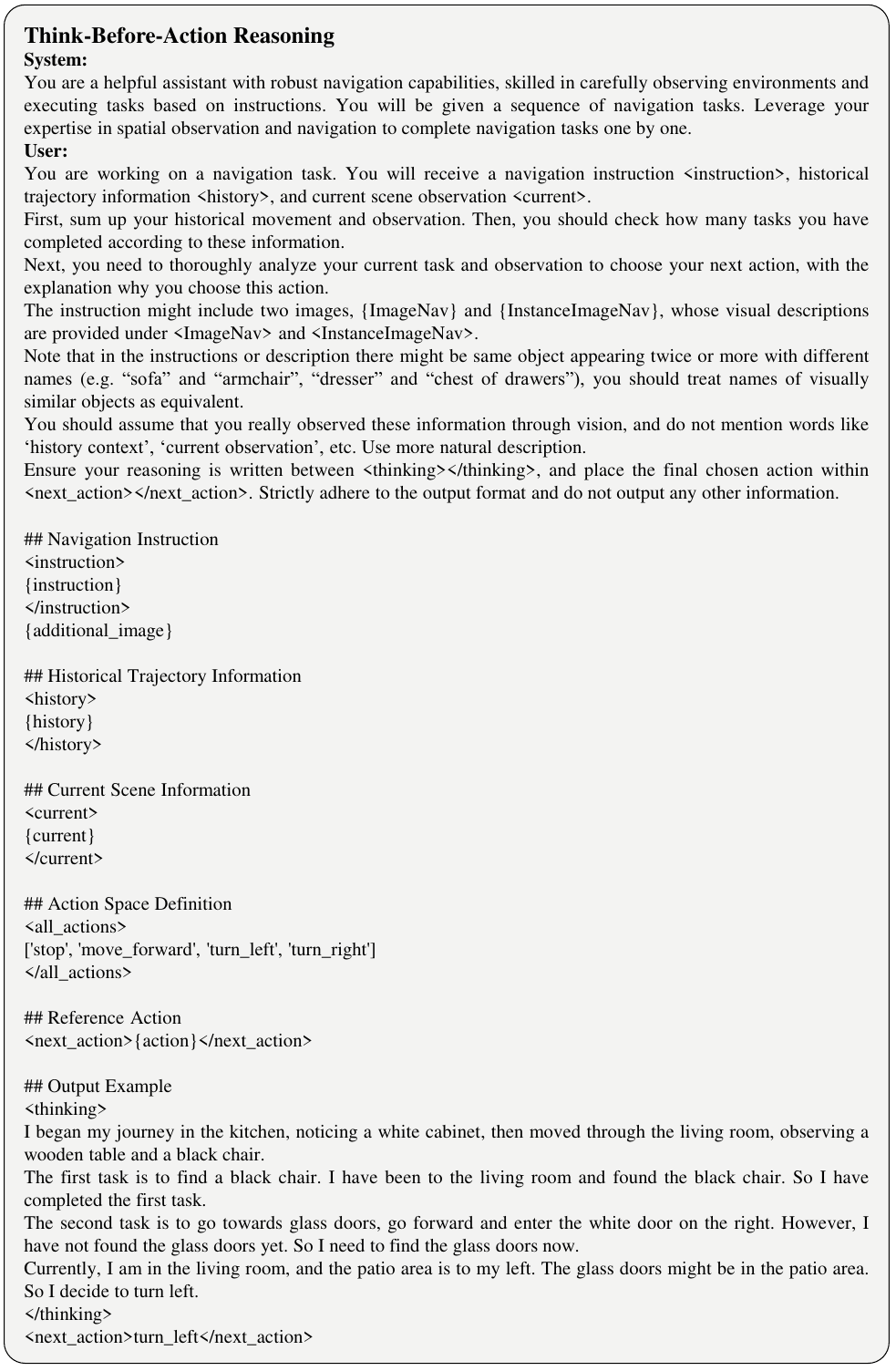}}
    \vspace{-1mm}
        \caption{Prompt for TBA Reasoning.}
    \vspace{-3mm}
        \label{fig:prompt_TBA}
    \end{center}
\end{figure}
\noindent\textbf{TBA Reasoning.}
The deep reasoning stage integrates current observations, historical trajectories, navigation instructions, and optional image goals into a structured prompt template for reasoning model to process. To align model outputs with ground truth actions while maintaining natural reasoning flow, we adopt a reference action embedding strategy with multiple rejection sampling. 
This approach provides target actions in the prompt context without explicit constraints, enabling the model to generate logically coherent reasoning rather than result-oriented justification. The detailed prompts are shown in Fig.~\ref{fig:prompt_TBA}.

\subsection{Data Analysis}
\label{append:bench-data}
\begin{figure}[t]
    \begin{center}
    \centerline{\includegraphics[width=0.6\linewidth]{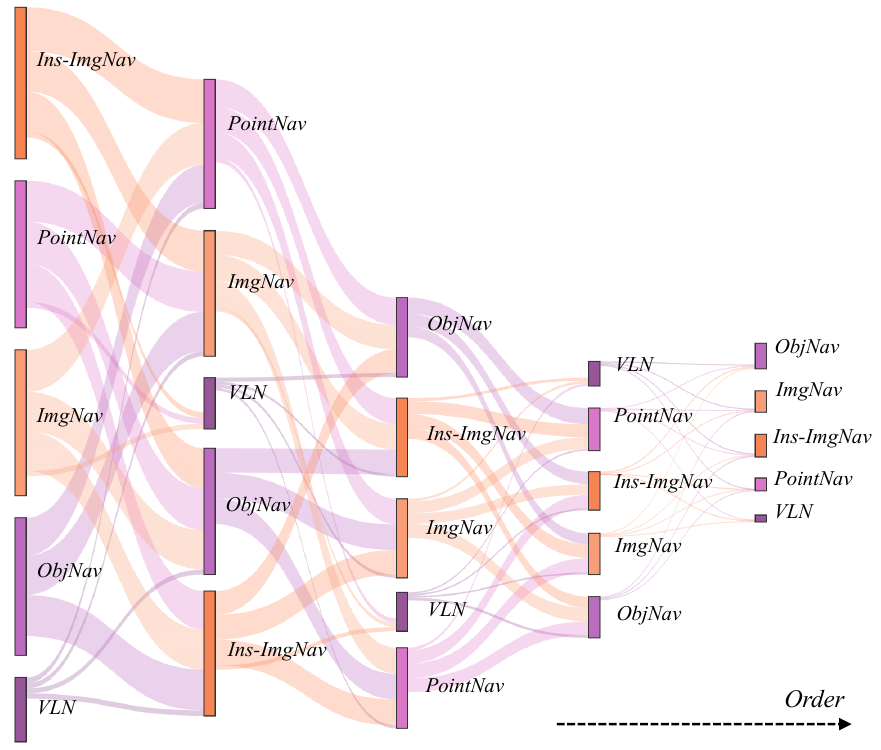}}
    \vspace{-1mm}
        \caption{Sankey diagram of capability distribution within instructions.}
    \vspace{-3mm}
        \label{fig:data_analysis_sangji}
    \end{center}
\end{figure}
\begin{figure}[t]
    \begin{center}
    \centerline{\includegraphics[width=1\linewidth]{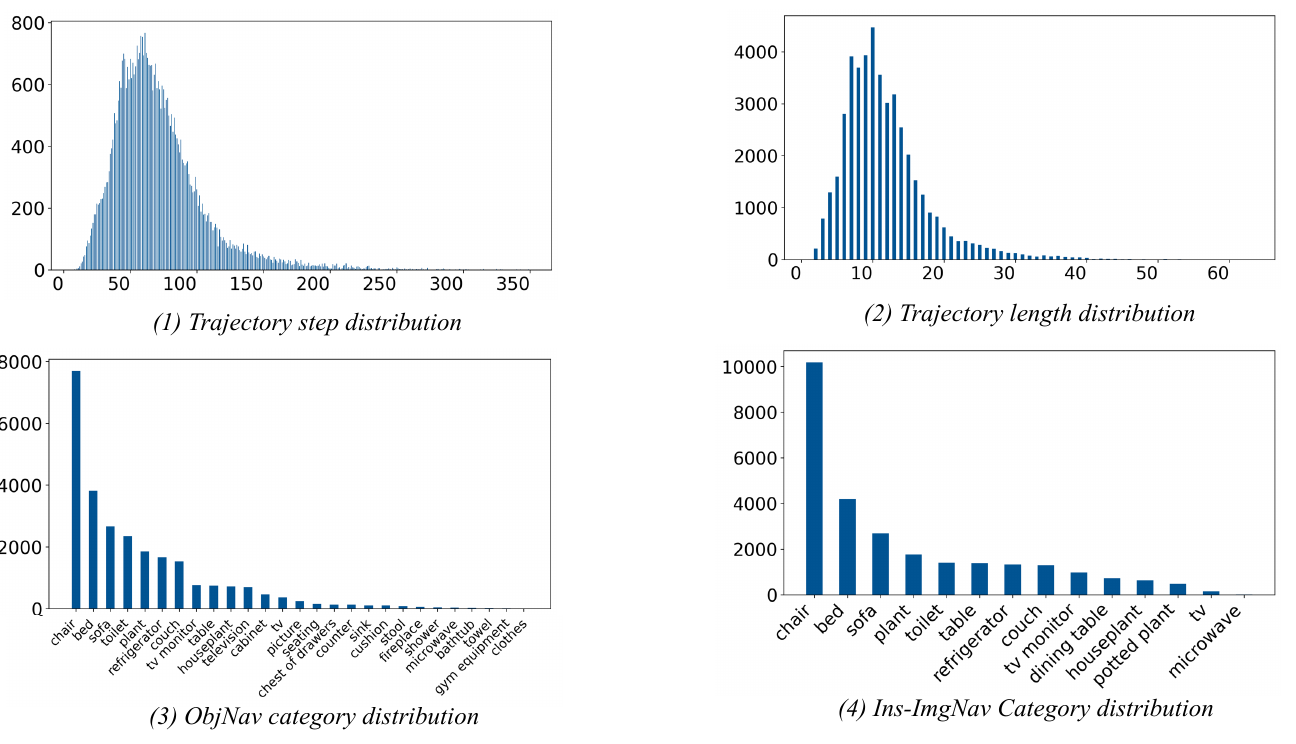}}
    \vspace{-1mm}
        \caption{Distribution of trajectory and category.}
    \vspace{-3mm}
        \label{fig:data_analysis_bar}
    \end{center}
\end{figure}
\noindent\textbf{Sankey Diagram.}
In Fig.~\ref{fig:data_analysis_sangji}, we visualize the data distribution by a Sankey diagram. 
Specifically, each instruction in OctoNav-Bench contains $1$ to $5$ types of navigation capabilities in order. Thus, each edge in the diagram represents an instruction, and edges with the same type of ability are grouped into a cluster. Note that the thickness of the cluster reflects the number of corresponding instructions.
which means the thicker the cluster, the greater the number of corresponding instructions. 
We can observe that the clusters become thinner toward the right, indicating that instructions containing $5$ distinct navigation capabilities (\ie, the hard instruction) are relatively rare, while instructions with $1$ to $4$ tasks appear more frequently.
Moreover, navigation capabilities appear in various orders across the instructions, confirming the diversity.
Additionally, we note that the frequency of the VLN task appears relatively lower compared to other tasks. This is primarily because VLN instructions tend to be longer in word count, even if fewer in number. As a result, the total textual proportion of VLN remains balanced within the dataset.

\noindent\textbf{Bar Charts.} In Fig.~\ref{fig:data_analysis_bar}, we illustrate the distribution of trajectory lengths and step counts, along with the distribution of target categories in ObjNav and Ins-ImgNav.
The trajectory distribution approximately follows a normal distribution. The average length is long, and the range of lengths is wider, reflecting the high complexity and the challenging nature of OctoNav-Bench. 
In addition, the distribution of target categories in both ObjNav and Ins-ImgNav presents a broad spectrum of navigation goals, confirming the generalization potential.

\begin{figure}
    \begin{center}
    \centerline{\includegraphics[width=1\linewidth]{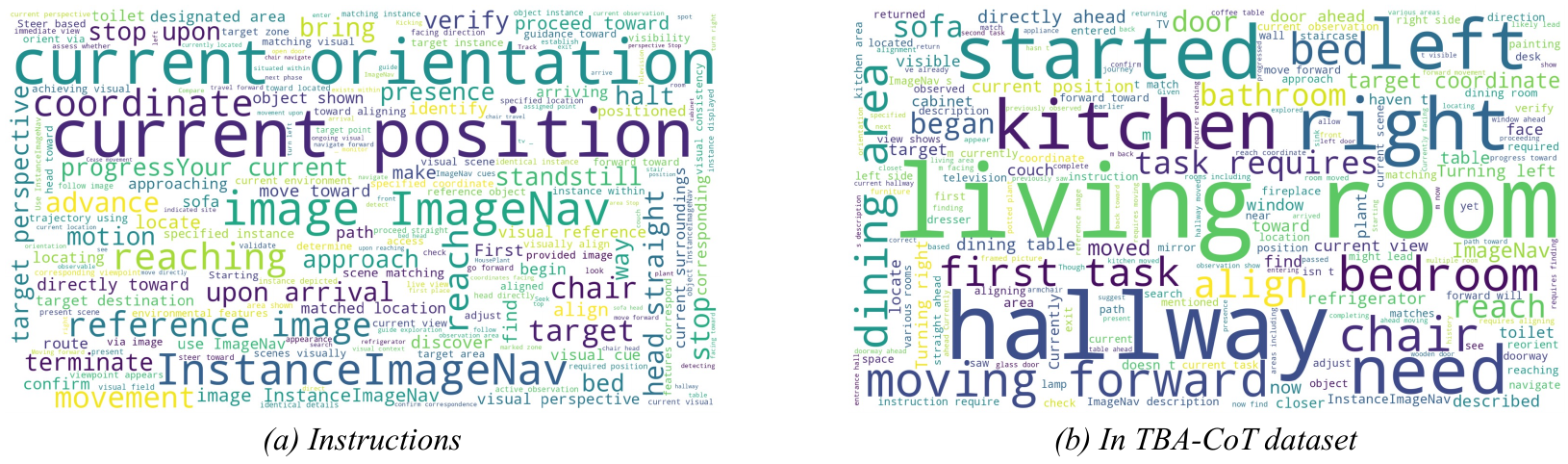}}
    \vspace{-1mm}
        \caption{Visualization of the word cloud.}
    \vspace{-3mm}
        \label{fig:cloud}
    \end{center}
\end{figure}

\noindent\textbf{Word Clouds.} As shown in Fig.~\ref{fig:cloud}, we analyze the word clouds for the instructions and TBA-CoT dataset of OctoNav-Bench. 
For instructions, 
there are plenty of words instructing to move like `proceed', `toward', `advance' and `approach'. The categories of ObjNav targets align with Fig.~\ref{fig:data_analysis_bar}, with chair, bed and sofa being the most frequently appearing words. For TBA-CoT dataset, the word cloud reveals several keywords in CoT. The room keywords like `living room', `hallway' and `kitchen' are emphasized, showing abundant location information is included in reasoning steps. In addition, `left' and `right' recurs frequently because of the spatial analysis of observations. Moreover, the words showing reasoning about task requirements and future actions are also mentioned repeatedly, such as `task requires', `need', `moving forward'.

\begin{figure}
    \begin{center}
    \centerline{\includegraphics[width=0.7\linewidth]{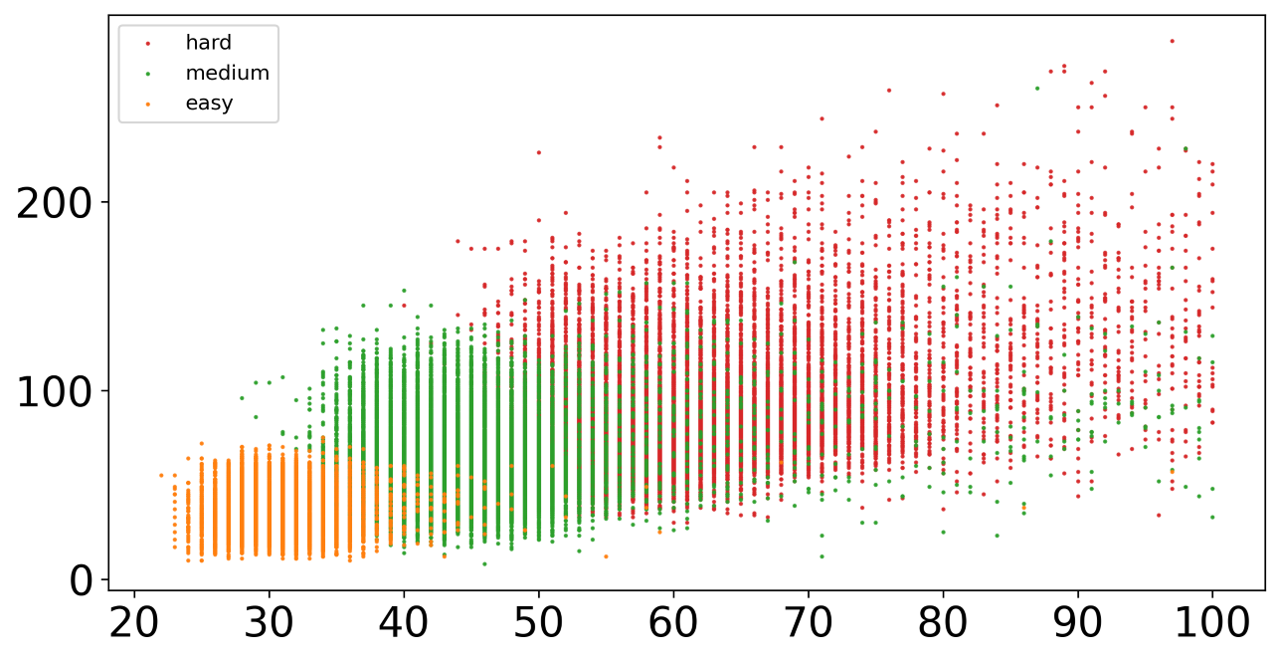}}
    \vspace{-1mm}
        \caption{Distribution of the task. The x-axis represents the number of words in the instruction, while the y-axis indicates the step length.}
    \vspace{-6mm}
        \label{fig:scatter}
    \end{center}
\end{figure}

\noindent\textbf{Scatter Chart.} Fig.~\ref{fig:scatter} shows the distribution of the task. The x-axis denotes the word number in the instruction, while the y-axis indicates the step length of the trajectory. To enhance chart clarity, data points with instruction exceeding $100$ words have been excluded. All data points are categorized into three groups: easy, medium, and hard, based on the number of capabilities involved. Easy tasks contain a single capability, medium tasks include two capabilities, while hard tasks have three or more tasks. In the figure, they are denoted by orange, green, and red, respectively. 
Collectively, task difficulty demonstrates correlation with both instruction length and step length, where longer instruction and trajectory correlate with increased task complexity. A roughly proportional relationship between instruction length and step length is also evident. The chart further reveals substantial diversity in instruction lengths within our dataset, particularly notable in the hard category where multiple capabilities interact to produce complex task requirements. This diversity highlights that our OctoNav-Bench is designed to cover a wide range of navigation situations.


\vspace{-1mm}
\subsection{Data Usage.}
The proposed OctoNav-Bench uses Habitat simulator~\cite{habitat19iccv,szot2021habitat,puig2023habitat}, MP3D~\cite{chang2017matterport3d}, HM3D~\cite{ramakrishnan2021hm3d}, Gibson~\cite{xia2018gibson}, and ProcTHOR~\cite{deitke2022️procthor} scenes, which are all public. Besides, the construction of many datasets~\cite{krantz2020beyond,khanna2024goat,song2024towards,zhu2021soon,yokoyama2024hm3d,krantz2023iterative} is also based on these scenes and simulators. 
All data are used in compliance with their licenses and terms of use.

\vspace{-1mm}
\subsection{Metrics Details}
\label{append:bench-metric}
We provide computation details about metrics for better understanding. Concretely, there are $N$ test samples in total. $i$-th sample contains $num_i$ sub-tasks and its $j$-th sub-task is defined as $sub_{i,j}$. Note that each $sub_{i,j}$ has a corresponding capability $cap_{i,j}$ of this sub-task.
Also, $S_{i,j}$ denotes the binary success indicator of this sub-task, and $OS_{i,j}$ denotes the binary indicator of oracle success. $Goal_{i,j}$ is the target point, $SD_{i,j}$ is the success distance. 

Following prior works~\cite{anderson2018evaluation, batra2020objectnav,zhang2024navid}, we set different success distances for each capability: $0.36$m for PointNav and ImgNav, $1$m for ObjNav and Ins-ImgNav, $3$m for VLN. Any location within the success distance of the target location is regarded as a successful area. $Area_{i,j}$ presents the successful area of the sub-task $sub_{i,j}$. Formally, it can be achieved via:
\begin{equation}
\small
Area_{i,j} = \{p| p \in P \land dis(Goal_{i,j},p) \le SD_{i,j}\},
\end{equation}
where $P$ denotes the whole navigable area. Then, let $Traj_{i,k}$ be the location after the $k$-th step of agent's trajectory in sample $i$. We obtain $S_{i,j}$ and $OS_{i,j}$ as follows:
\begin{equation}
\small
OS_{i,j} = \exists k, \text{ s. t. } Traj_{i,k} \in Area_{i,j},
\end{equation}
\begin{equation}
\small
S_{i,j} = \exists (k_1, \dots,k_j), \text{ s. t. } (k_1 < \dots < k_j)\land (Traj_{i,k_1}\in Area_{i,1})\land \dots \land (Traj_{i,k_j}\in Area_{i,j}).
\end{equation}
Let $L_{i,j}$ be the shortest path length from the starting point satisfying $S_{i,j}=1$. 
Note that $L_{i,0}=0$. 
Next, we define $l_{i,j}$ as the local shortest path of $sub_{i,j}$, which equals to $L_{i,j} - L_{i,j-1}$. Similarly, let $TL_{i,j}$ be the length of the trajectory actually taken by the agent, which meets $S_{i,j}=1$. Note that $TL_{i,0}=0$. 
And $tl_{i,j}$ is defined as the local trajectory length of $sub_{i,j}$, which equals to $TL_{i,j} - TL_{i,j-1}$. 
If $sub_{i,j}$ is not successful, both $TL_{i,j}$ and $tl_{i,j}$ are considered as $+\infty$. 
Thus, for each capability $c\in[\text{Ins-ImgNav},\text{ImgNav},\text{PointNav},\text{ObjNav},\text{VLN}]$, the evaluation metrics can be calculated as follows:
\begin{equation}
\small
SR_{c} = \frac{1}{\sum\limits_{i=1}^N \sum\limits_{j=1}^{num_i}[cap_{i,j} = c]}\sum\limits_{i=1}^N \sum\limits_{j=1}^{num_i}S_{i,j}[cap_{i,j} = c],
\end{equation}
\begin{equation}
\small
OSR_{c} = \frac{1}{\sum\limits_{i=1}^N \sum\limits_{j=1}^{num_i}[cap_{i,j} = c]}\sum\limits_{i=1}^N \sum\limits_{j=1}^{num_i}OS_{i,j}[cap_{i,j} = c],
\end{equation}
\begin{equation}
\small
SPL_{c} = \frac{1}{\sum\limits_{i=1}^N \sum\limits_{j=1}^{num_i}[cap_{i,j} = c]}\sum\limits_{i=1}^N \sum\limits_{j=1}^{num_i}S_{i,j}\frac{l_{i,j}}{max(l_{i,j},tl_{i,j})}[cap_{i,j} = c].
\end{equation}
And overall evaluation metrics can be calculated as follows:
\begin{equation}
\small
SR_{Overall} = \frac{1}{N}\sum\limits_{i=1}^N S_{i,num_i},
\end{equation}
\begin{equation}
\small
OSR_{Overall}= \frac{1}{N}\sum\limits_{i=1}^N OS_{i,num_i},
\end{equation}
\begin{equation}
\small
SPL_{Overall} = \frac{1}{N}\sum\limits_{i=1}^N S_{i,num_i}\frac{L_{i,num_i}}{max(L_{i,num_i},TL_{i,num_i})}
\end{equation}

\begin{figure}[t]
\begin{center}
  \includegraphics[width=0.25\linewidth]{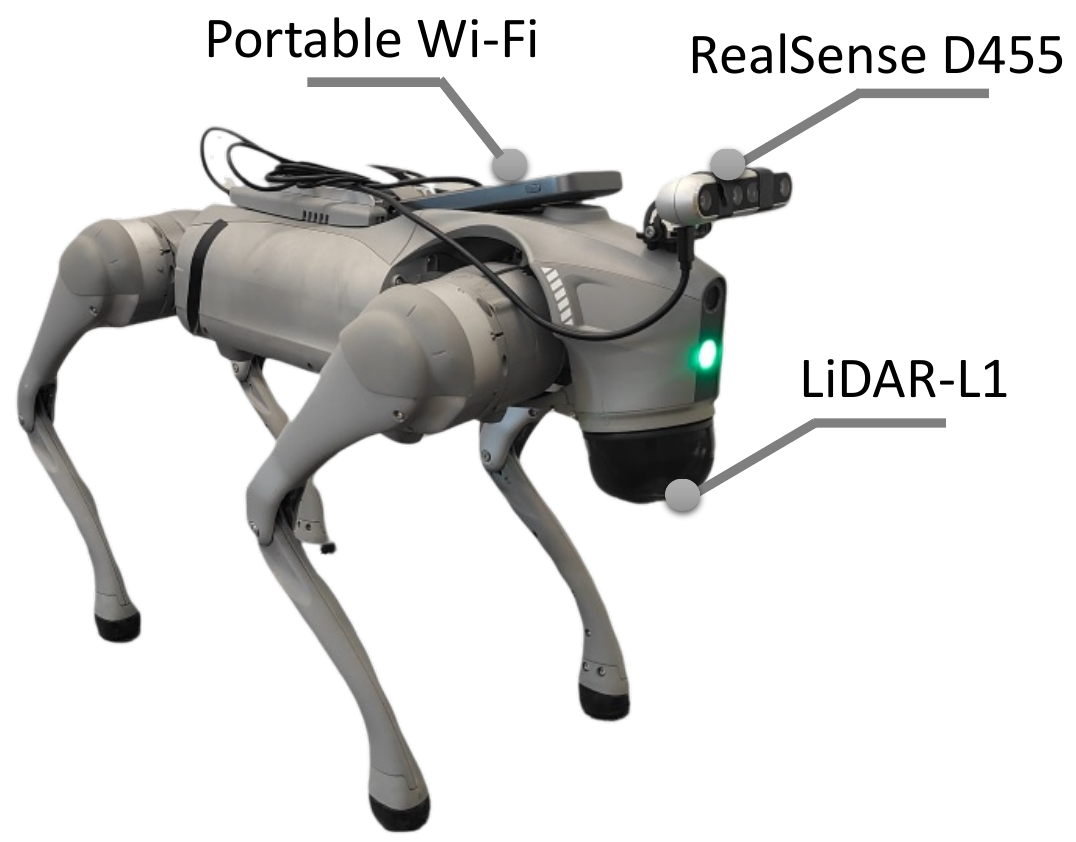}
\end{center}
   \caption{We use Unitree GO2, and mount RealSense D455, a portable Wi-Fi and a LiDAR-L1. Note that, our model only takes RGB frames as input. The portable Wi-Fi is used for communication with the remote server and the Lidar is used for the local controller API of Unitree Dog.}
\label{fig:robot_setup}
\vspace{-1mm}
\end{figure}

\section{Appendix: Real-World Deployment}
\label{append-real}
\subsection{Robot Setup}
OctoNav-R1 is tested on a robotic dog (Unitree GO2) mounted with a RealSense D455 camera on its head (We provide a visualization at Fig.~\ref{fig:robot_setup}). Here, we only use the RGB frames with a resolution of $640\times480$ in the setting of  $90^\circ$ HFOV. We also mount a portable Wi-Fi at the back of the robot dog, which is used to communicate with the remote server (send captured images and receive commands). Unitree GO2 is integrated with a LiDAR-L1, which is only used for local motion planning for Unitree's API. 

\subsection{Deployment Details}
Our model is deployed on a remote server equipped with an NVIDIA 4090 GPU. During navigation, the server receives navigation instructions and images captured by the robotic dog through the Internet. To ensure efficient communication, the images are compressed before transmission. After processing the newly captured images, the model generates action commands and sends them to the robotic dog. Upon receiving these commands, the robotic dog executes the actions using a local motion planning model (specifically, the off-the-shelf model provided by Unitree Dog).

\subsection{Visualization in the Real-World}
\begin{figure}[t]
    \begin{center}
    \centerline{\includegraphics[width=0.7\linewidth]{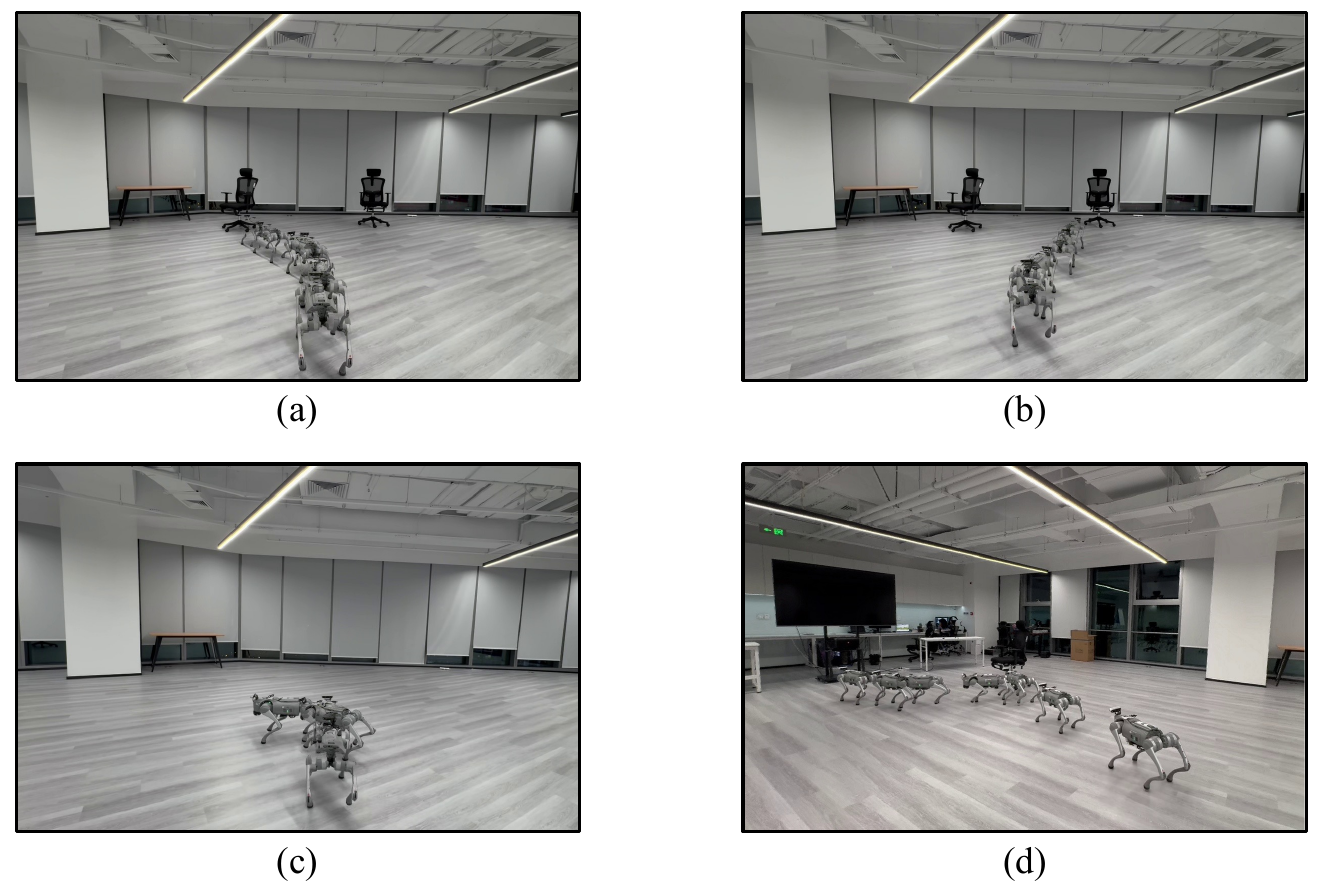}}
        \caption{{Visualization in real world.}}
    \vspace{-6mm}
        \label{fig:vis_OctoNav_real}
    \end{center}
\end{figure}
\begin{figure}
    \begin{center}
    \centerline{\includegraphics[width=0.8\linewidth]{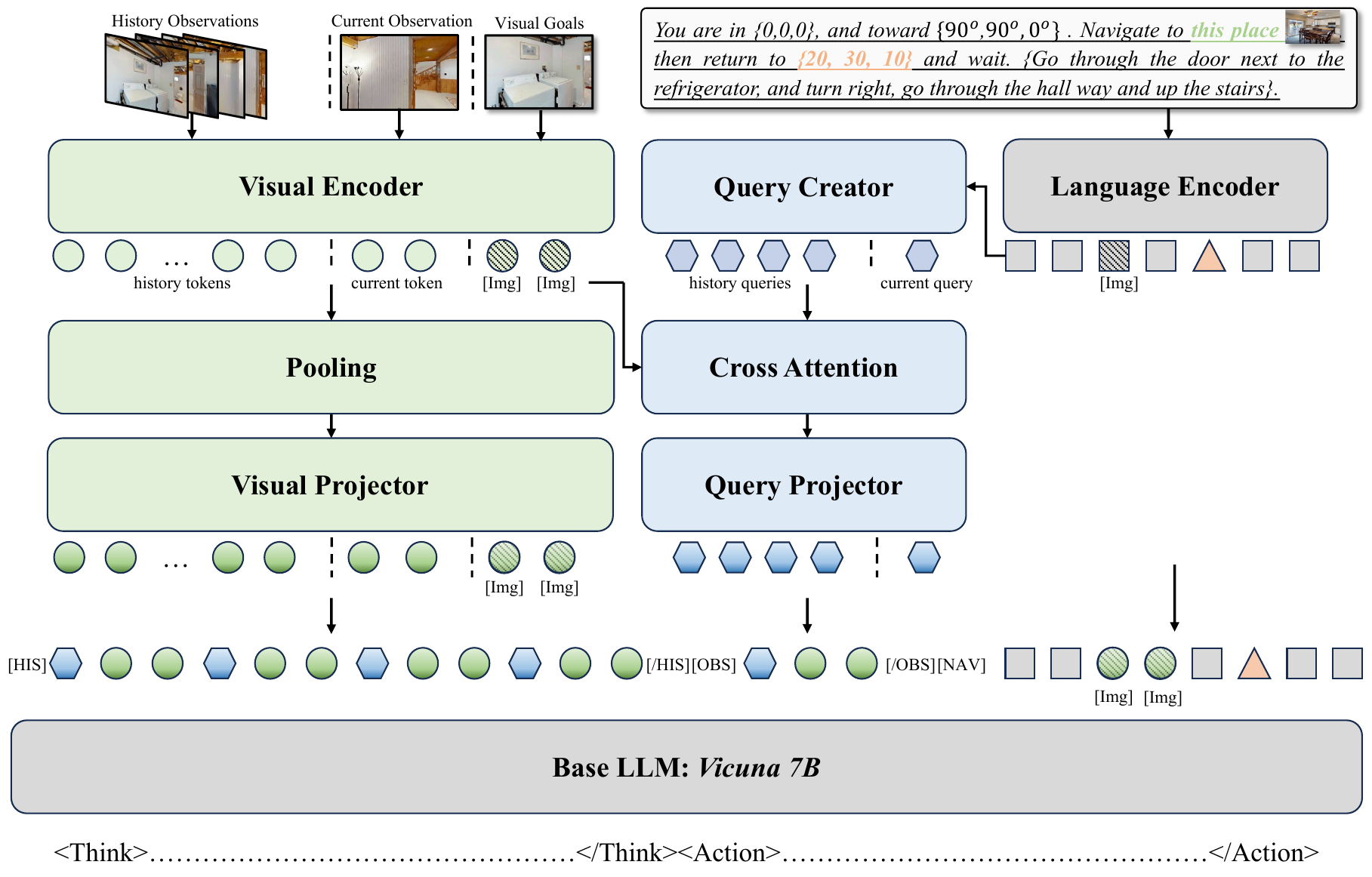}}
        \caption{{The model architecture of OctoNav-R1. Best viewed in color.}}
    \vspace{-5mm}
        \label{fig:model}
    \end{center}
\end{figure}
We present visualizations in the real world in Fig.~\ref{fig:vis_OctoNav_real}. In Fig.~\ref{fig:vis_OctoNav_real}(a) and Fig.~\ref{fig:vis_OctoNav_real}(b), the robot is required to navigate to the left chair and the right chair respectively. As shown in the visulizations, the robot correctly distinguish left and right. Besides, the robot in Fig.~\ref{fig:vis_OctoNav_real}(b) successfully navigates to the required point. All these demonstrate that OctoNav-R1 can accurately recognize the spatial relationship. In Fig.~\ref{fig:vis_OctoNav_real}(d) is required to find a chair first, and then navigate to the tv shown in an image. As shown in Fig.~\ref{fig:vis_OctoNav_real}(d), our robot successfully finishes this multi-stage task, indicating the ability of multi-modal cognition and task planning. We observe preliminary sim-to-real transfer ability without real-world fine-tuning. It further confirms the OctoNav-Bench and OctoNav-R1.

\section{Appendix: More Technical Details of OctoNav-R1}
\label{append:method}
\subsection{Model Architecture}
\label{append:architecture}
Our OctoNav-R1 is built upon LLaMA-VID~\cite{li2024llama}. 
Further, we propose task-specific designs on top of it, making it capable of receiving multi-modal instructions and producing low-level actions in an end-to-end manner.
The detailed model architecture of OctoNav-R1 is shown in Fig.~\ref{fig:model}. Firstly, all images are encoded by the visual encoder $\mathcal E_V$:
\begin{equation}
\mathbf E_{h}^i=\mathcal E_V(\mathcal V_h^i),\mathbf E_{c}=\mathcal E_V(\mathcal V_c),\mathbf E_{g}=\mathcal E_V(\mathcal V_g),
\end{equation}
where $\mathcal V_h^i$ is the $i$-th historical observation, $\mathcal V_c$ is the current observation, $\mathcal V_g$ is the goal image for ImgNav or Ins-ImgNav tasks. These image features are further sent into the pooling layer and the visual projector $\mathcal P_V$ to align with the LLM embedding space:
\begin{equation}
\mathbf E{'}_{h}^{i}=\mathcal P_{V}(\text{Pool}(\mathbf E_{h}^i)),\mathbf E'_{c}=\mathcal P_{V}(\text{Pool}(\mathbf E_c)),\mathbf E'_{g}=\mathcal P_{V}(\text{Pool}(\mathbf E_g)).
\end{equation}
Meanwhile, the instruction $\mathcal I$ is embedded by the language encoder $\mathcal E_{L}$:
\begin{equation}
\mathbf I = \mathcal E_{L}(\mathcal I).
\end{equation}
Besides, for each observation, we generate a query token for it using the Q-Former style query generator $\mathcal G_{Q}$:
\begin{equation}
\mathbf Q_h^i=\mathcal G_{Q}(\mathbf E_h^i,\mathbf I),\mathbf Q_c=\mathcal G_{Q}(\mathbf E_c,\mathbf I).
\end{equation}
These tokens are then aligned to the LLM embedding space via cross attention and query projector $\mathcal P_Q$:
\begin{equation}
\mathbf Q{'}_h^i=\mathcal P_{Q}(\text{CA}(\mathbf E_h^i,\mathbf Q_h^i)),\mathbf Q'_c=\mathcal P_{Q}(\text{CA}(\mathbf E_c,\mathbf Q_c)),
\end{equation}
where $\text{CA}(\cdot)$ is the cross attention function:
\begin{equation}
\text{CA}(\mathbf E,\mathbf Q)=\text{Pool}(\text{Softmax}(\mathbf Q\mathbf E^T)\mathbf E).
\end{equation}

\begin{figure}[H]
    \begin{center}
    \centerline{\includegraphics[width=0.8\linewidth]{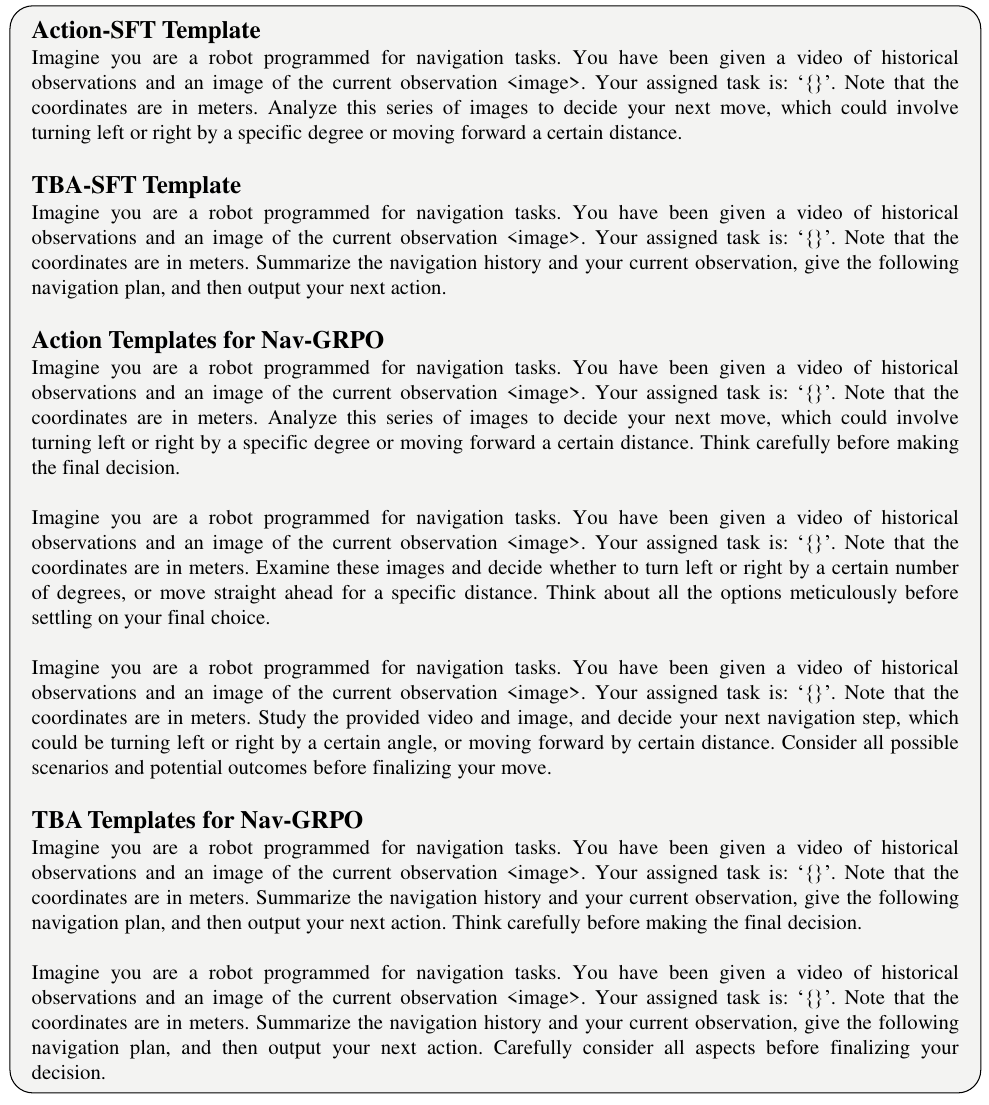}}
    \vspace{-1mm}
        \caption{Prompt for Training.}
    \vspace{-3mm}
        \label{fig:prompt_for_training}
    \end{center}
\end{figure}

Then each query token embedding is inserted to the head of the corresponding image embeddings. 
The embedding of the placeholder (\eg, \{ImageNav\} and \{InstanceImageNav\}) is replaced by the goal image embedding $\mathbf E'_{g}$, as shown in Fig.~\ref{fig:model}. In practice, we utilize EVA-CLIP~\cite{sun2023eva} as the visual encoder, BERT~\cite{devlin2019bert} as the Language Encoder, and Vicuna-7B~\cite{vicuna2023} as the base LLM.

\subsection{Implementation Details.}
\label{append:implementation}
For the Action-SFT stage, we train the model for nearly $10k$ steps via the instruction-trajectory pairs within OctoNav-Bench. 
Then we train the model on the TBA-CoT dataset of OctoNav-Bench for nearly $6k$ steps to enable the model's TBA ability. 
For the GRPO training stage, $N_{GRPO}$=$2,000$, and we train the model for $1,000$ steps with $\varepsilon=0.2$ and $\beta=0.0001$.
All trainings are finished on $8$ A800 40G GPUs with a learning rate of $2e-5$, and the batch size for a single GPU is $2$.
For the online reinforcement learning, we train the model with a learning rate of $2e-6$ on a single A800 for $500$ steps. The discount factor $\gamma=1$, $d'=25cm$, and the warm-up stage contains $100$ steps. 
To enable efficient fine-tuning, we employ LoRA~\cite{hu2021lora} technique across all training stages. The prompts used in training are presented in Fig.~\ref{fig:prompt_for_training}.
\subsection{Baseline Methods}
\label{append:baseline}
We provide more details of the baselines for comparison and how they are modified to evaluate on OctoNav-Bench.

\noindent\textbf{MLLMs as Agent.} We select representative open-source MLLM for baseline: Qwen-VL~\cite{bai2025qwen2}, Video-LLaVA~\cite{lin2024video} and LLaVA-NeXT-Video~\cite{zhang2024llavanextvideo}. For Qwen-VL, we combine historical observation images, current observation image and image goals into a single large picture, with a caption on top of each image. Then, the picture and the instruction are fed into MLLM to produce action. 
For Video-LLaVA and LLaVA-NeXT-Video, the historical observation images are sent to model through video, while the current observation image and image goals are sent through a combined picture.

\noindent\textbf{Navigation Model for Discrete Environments.} NaviLLM~\cite{zheng2024towards} pioneers as a universal framework for embodied navigation scenarios, tailoring LLMs to address diverse task requirements through schema-based instruction. NavGPT-2\cite{zhou2024navgpt2} leverages the VLM framework from InstructBLIP~\cite{dai2023instructblip}, enhanced with multi-image perception capabilities to optimize its performance for VLN tasks. As both models are developed in discrete environments, we finetune them to fit in our continuous environment. 

For NaviLLM, we first change the output head from waypoint-selection to action-selection, aligning with continuous environment setting. Next, we slightly modify the instruction template, adding current observation, ImgNav goal and Ins-ImgNav goal tokens into input while keeping the original history image tokens unchanged. The corresponding images are sent into visual encoder and embedded as visual input. After modification, we fine-tune the LLM backbone using the LoRA~\cite{hu2021lora} method.

For NavGPT-2, we apply its first training stage (\ie, LLM finetuning) on the OctoNav-Bench, thus enable the LLM to output actions needed in the continuous environments. Particularly, we put text "ImageNav" and "InstanceImageNav" on the corresponding goal images, send all images to the visual encoder, and stack all image embeddings as the visual input. In inference, the navigation action is directly extracted from the LLM output text.  

\noindent\textbf{Navigation Model for Continuous Environments.} NaVid~\cite{zhang2024navid} and Uni-NaVid~\cite{zhang2025uninavid} are two video-based Vision-Language Models for VLN, developed in continuous environment like ours. Uni-NaVid is a follow-up to NaVid, with much more training data and stronger navigation ability. We firstly add image goal embeddings into the models, then evaluate them before and after finetuning on the OctaNav-Bench.

\section{Appendix: More Experimental Results}
\subsection{Visualization on OctoNav-Bench}

In this section, we provide more visualizations of navigation trajectories and thinking processes of OctoNav-R1.

\begin{figure}[H]
    \begin{center}
    \centerline{\includegraphics[width=1\linewidth]{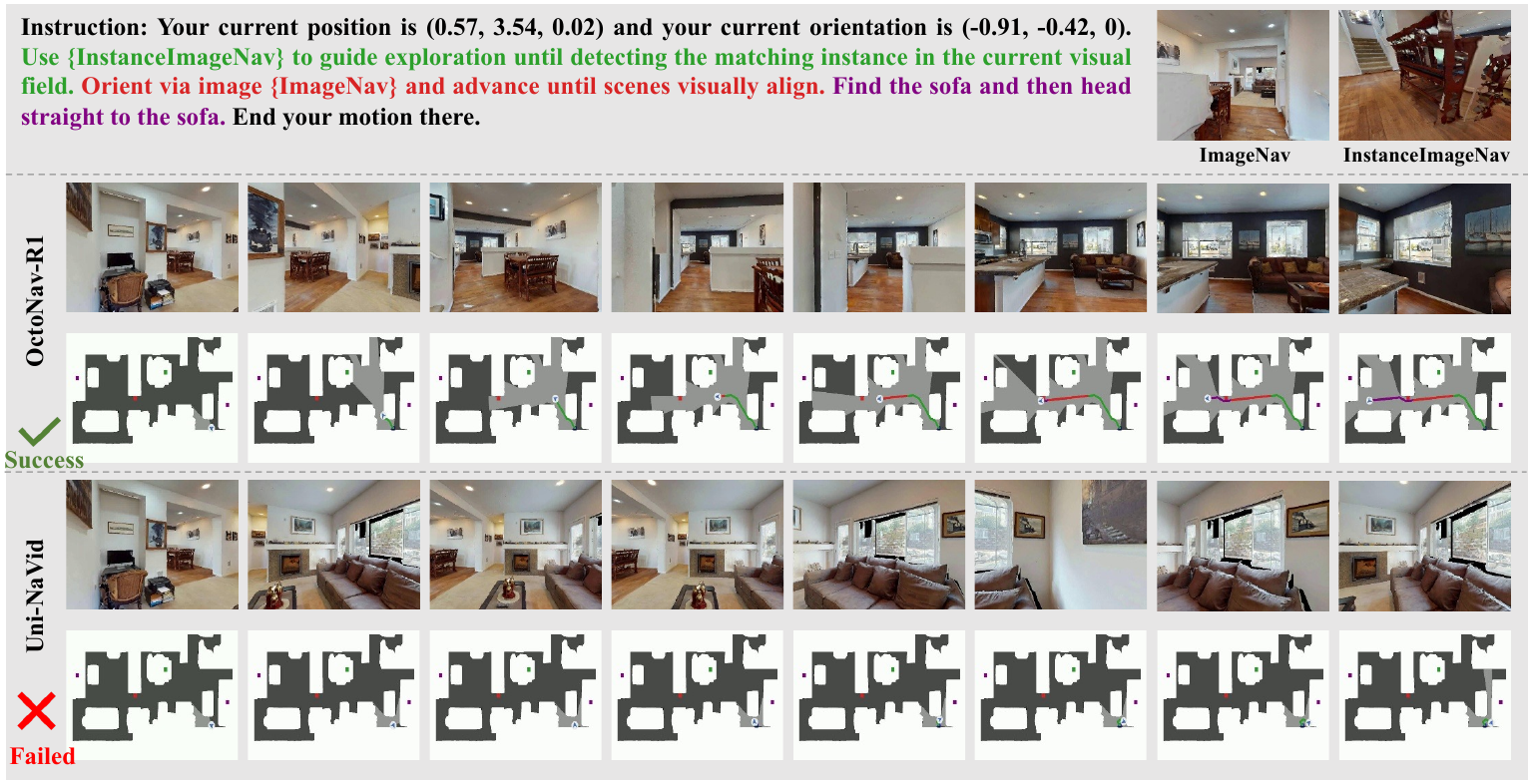}}
        \caption{{Visualization on OctoNav-Bench.}}
    \vspace{-3mm}
        \label{fig:vis_OctoNav_0}
    \end{center}
\end{figure}

In Fig.~\ref{fig:vis_OctoNav_0}, the agent is required to reach the scene in \{InstanceImageNav\} and \{ImageNav\} subsequently, and finally find the sofa. For OctoNav-R1, as the chairs in \{InstanceImageNav\} are already in the field of view, the agent directly walks towards these chairs. After finishing this task, the agent notices that the hallway in \{ImageNav\} is on the left side. In this way, the agents turns left and walks through the hallway, where the sofa emerges. Then the agent walks close to the sofa and stops. Such a trajectory shows that OctoNav-R1 can correctly understand the sequence of tasks, determine whether the current task is completed, and switch to the next task. In contrast, the agent of Uni-NaVid only stays beside the sofa, as the agent ignores the first two tasks and only focuses on the last one.

\begin{figure}[H]
    \begin{center}
    \centerline{\includegraphics[width=1\linewidth]{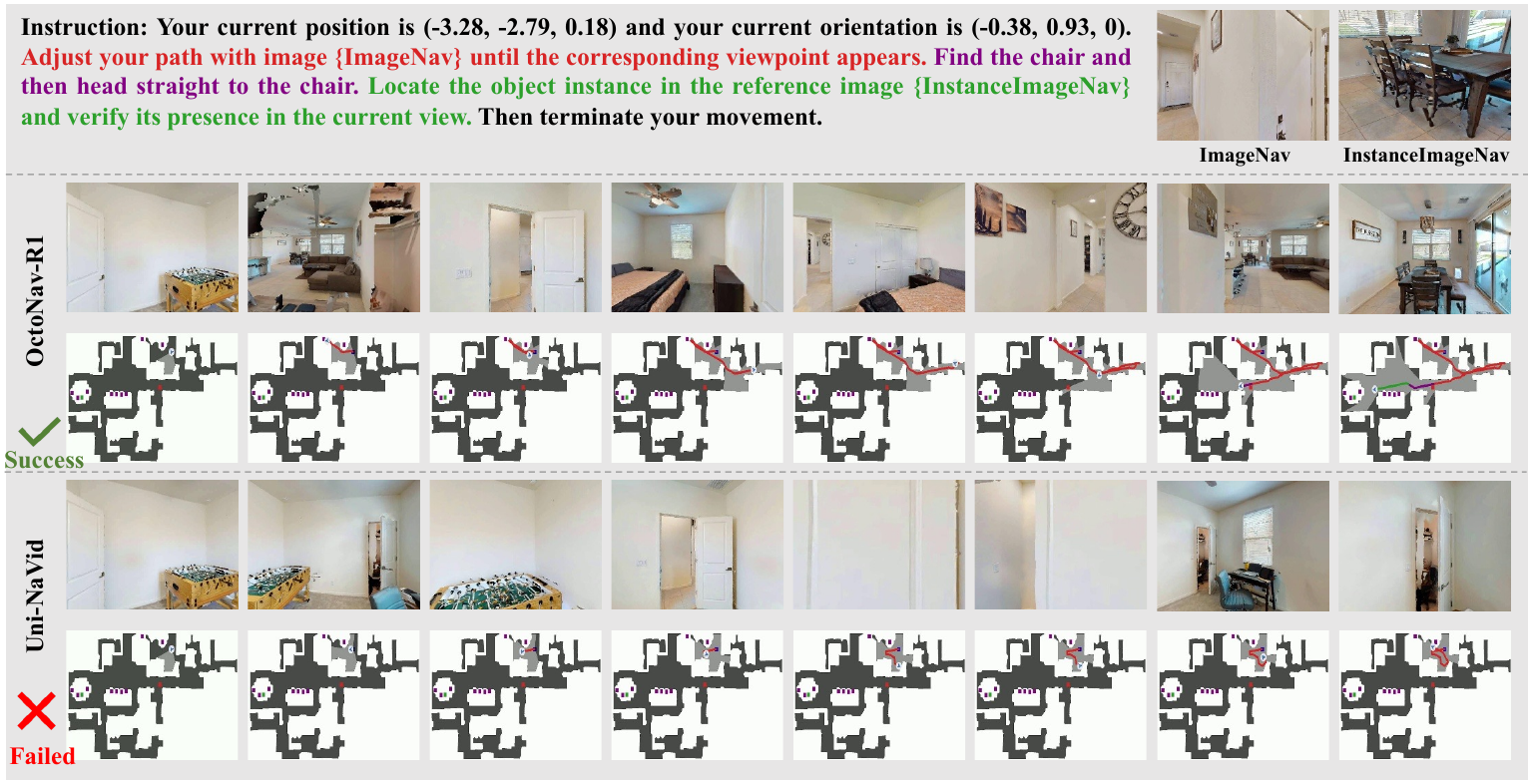}}
        \caption{{Visualization on OctoNav-Bench.}}
    \vspace{-3mm}
        \label{fig:vis_OctoNav_1}
    \end{center}
\end{figure}

In Fig.~\ref{fig:vis_OctoNav_1}, the agent is required to reach the scene in \{ImageNav\}. As the \{ImageNav\} shows a scene in the hallway, the agent of OctoNav-R1 firstly explores the hallway in the right area. After confirming that the required scene is not located in the right hallway, the agent turns back to explore the hallway at left, where it successfully reaches the required scene. Such a process indicates that OctoNav-R1 can precisely understand the contents in the goal images, and flexibly adjusts the navigation plan when the target cannot be found. In contrast, Uni-NaVid fails to understand the image content and hovers at the starting point.

\begin{figure}[H]
    \begin{center}
    \centerline{\includegraphics[width=1\linewidth]{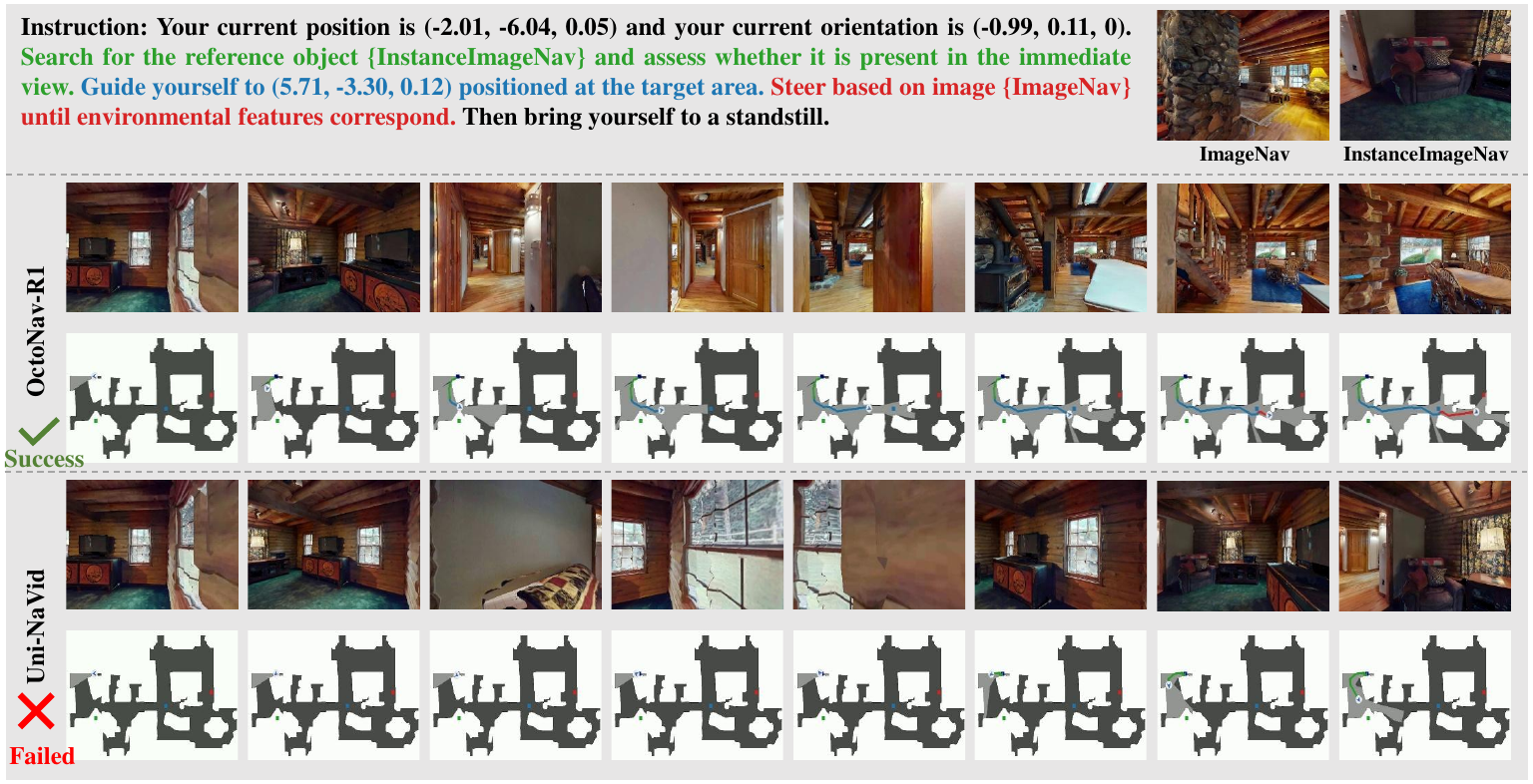}}
        \caption{{Visualization on OctoNav-Bench.}}
    \vspace{-3mm}
        \label{fig:vis_OctoNav_2}
    \end{center}
\end{figure}

In Fig.~\ref{fig:vis_OctoNav_2}, our agent firstly finds the sofa shown in \{ImageNav\}. Then the agent directly navigates to the target point. Finally, the agent successfully reaches the scene shown in \{InstanceImageNav\}. Note that each target in this instruction is either point or image, showing the OctoNav-R1's ability to complete multi-modal targets. In contrast, Uni-NaVid only completes the first task.

\begin{figure}[H]
    \begin{center}
    \centerline{\includegraphics[width=1\linewidth]{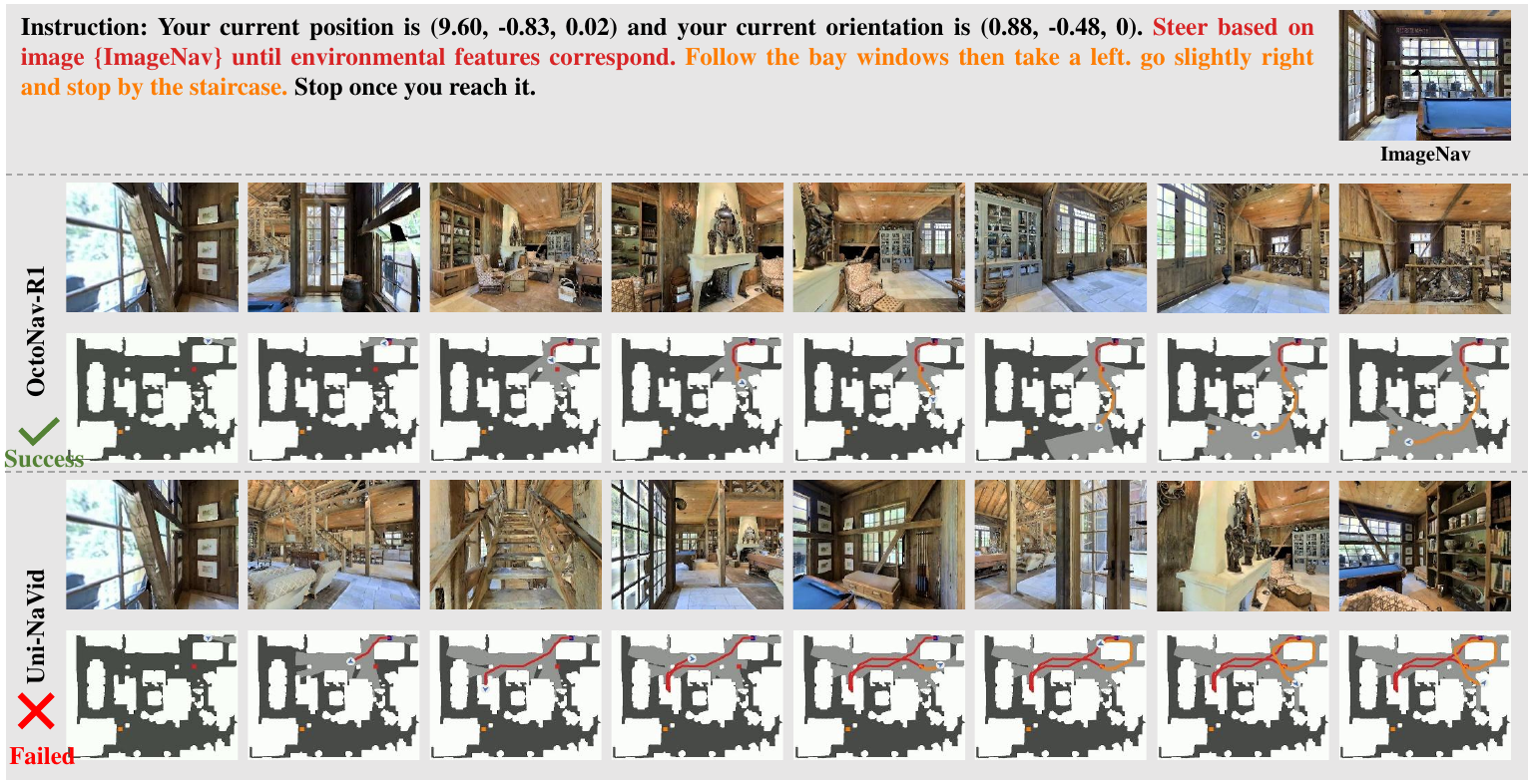}}
        \caption{{Visualization on OctoNav-Bench.}}
    \vspace{-3mm}
        \label{fig:vis_OctoNav_3}
    \end{center}
\end{figure}

In Fig.~\ref{fig:vis_OctoNav_3}, our agent first navigates to the scene displayed in \{ImageNav\}, and then follows a VLN instruction. It could be seen from the images that our agent can precisely obey the VLN instruction step-by-step, approaches the window first, and eventually reaches the staircase. Such a process demonstrates the OctoNav-R1's capability in completing VLN tasks. In contrast, Uni-NaVid agent tends to randomly wander randomly when faced with the complex instruction.

\begin{figure}[H]
    \begin{center}
    \centerline{\includegraphics[width=1\linewidth]{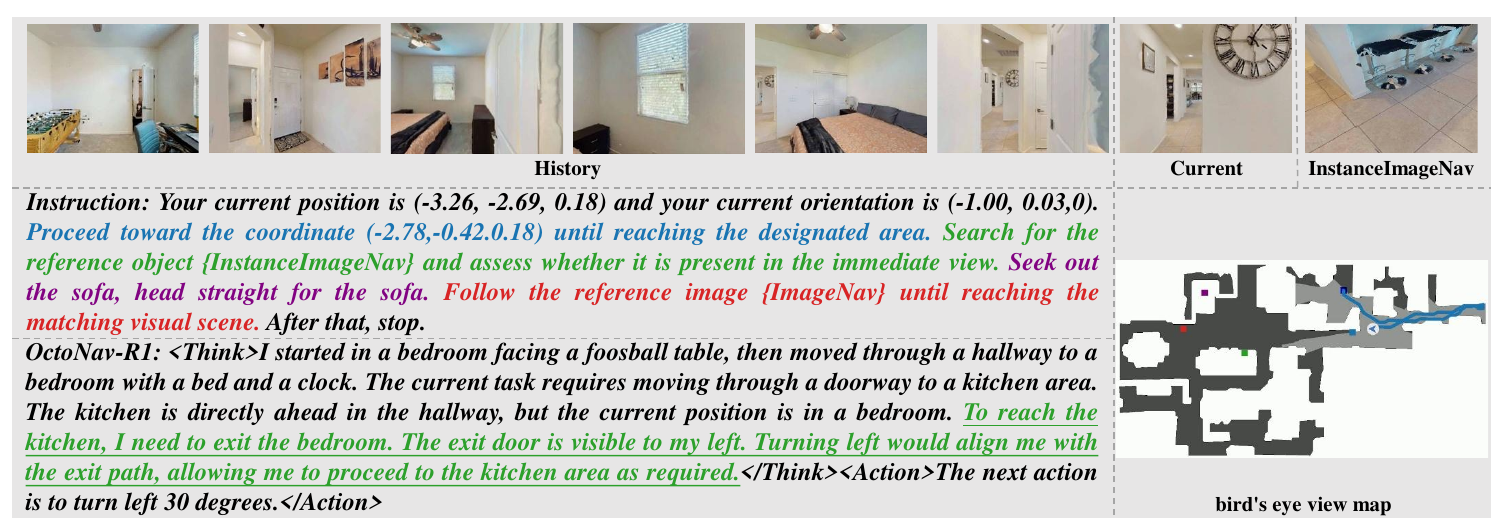}}
        \caption{{Visualization of think-before-actioin in a trajectory.}}
    \vspace{-3mm}
        \label{fig:vis_thinking_1}
    \end{center}
\end{figure}

In Fig.~\ref{fig:vis_thinking_1}, the agent is currently required to navigate to the kitchen area in the \{InstanceImageNav\}. Noticing the kitchen is ahead in the hall way, the agent decides to exit the current area. As the exit door is visible on the left of the current view, the agent turns left to align itself with the exit path. Such a process demonstrates that OctoNav-R1 can accurately understand the meaning of each image and extract useful information from different images.

\begin{figure}[H]
    \begin{center}
    \centerline{\includegraphics[width=1\linewidth]{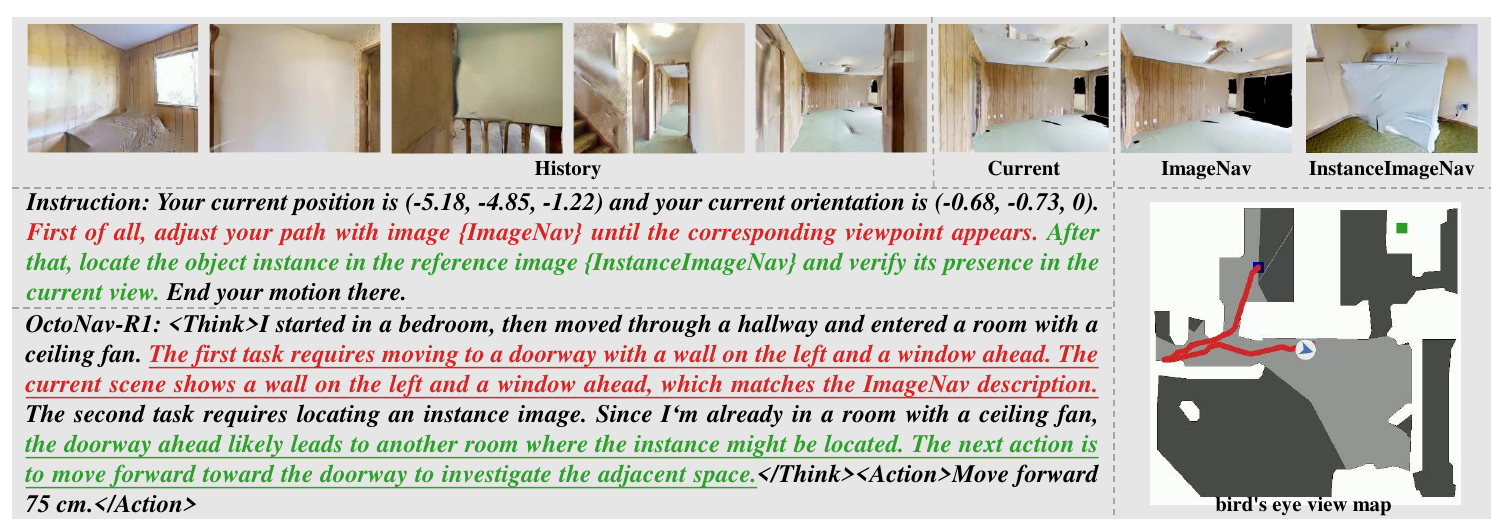}}
        \caption{{Visualization of think-before-actioin in a trajectory.}}
    \vspace{-3mm}
        \label{fig:vis_thinking_2}
    \end{center}
\end{figure}

In Fig.~\ref{fig:vis_thinking_2}, the agent currently reaches an empty room. By comparing the current observation with \{ImageNav\}, the agent correctly concludes that the first task has been completed, and continues to plan for the second task. As the target in \{InstanceImageNav\} does not appear in the current scene, 
\begin{figure}[H]
    \begin{center}
    \centerline{\includegraphics[width=1\linewidth]{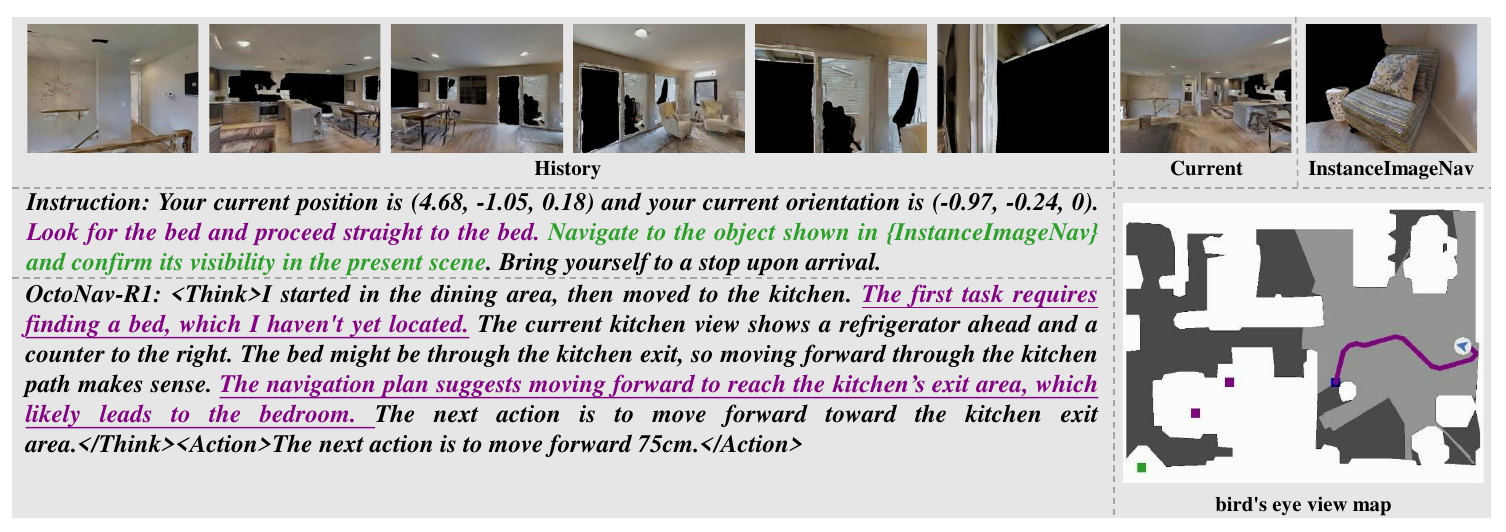}}
        \caption{{Visualization of think-before-actioin in a trajectory.}}
    \vspace{-3mm}
        \label{fig:vis_thinking_3}
    \end{center}
\end{figure}

the agent decides to explore the adjacent space for more clues. Such a behaviour shows that OctoNav-R1 can effectively make the navigation plan when faced with difficulties.

In Fig.~\ref{fig:vis_thinking_3}, the agent first navigates to the wrong direction which is opposite the bed target. Thankfully, through the TBA scheme, the agent realizes that the bed target is not located in the current area. Subsequently, the agent decides to leave the current area, and moves to the bedroom where the bed target is more likely to be. Such a process shows that OctoNav-R1 has the ability to correct errors in the navigation process.

\begin{table}[H]
\centering
\resizebox{\textwidth}{!}{
{
\renewcommand{\arraystretch}{1.3}
\setlength{\tabcolsep}{5pt}
\begin{tabular}{>{\raggedleft\arraybackslash}p{2.6cm}||ccc||ccc||ccc||ccc||ccc||ccc}
\hline \thickhline

 \rowcolor[HTML]{f8f9fa} \multicolumn{1}{c||}{}& \multicolumn{3}{c||}{Overall} & \multicolumn{3}{c||}{Ins-ImgNav} & \multicolumn{3}{c||}{ImgNav} & \multicolumn{3}{c||}{PointNav} & \multicolumn{3}{c||}{ObjNav} & \multicolumn{3}{c}{VLN}\\
\rowcolor[HTML]{f8f9fa} \multicolumn{1}{c||}{\multirow{-2}{*}{Template Type}} & SR& SPL & OSR & SR& SPL & OSR & SR& SPL & OSR & SR& SPL & OSR & SR& SPL & OSR & SR& SPL & OSR\\
\hline
\hline
Single Template & 15.80 & 10.93 & \textbf{25.40} & 27.02 & 19.58 & 33.06 & 17.36 & 12.30 & 19.42 & 21.12 & 14.71 & 23.11 & 47.13 & 34.57 & \textbf{63.52} & 31.43 & 28.23 & \textbf{48.57}\\
Diverse Templates & \textbf{17.00} & \textbf{12.04} & 25.00 & \textbf{29.44} & \textbf{20.73} & \textbf{33.47} & \textbf{19.42} & \textbf{14.27} & \textbf{22.73} & \textbf{21.51} & \textbf{16.37} & \textbf{24.70} & \textbf{48.77} & \textbf{35.63} & 62.70 & \textbf{34.29} & \textbf{29.14} & \textbf{48.57}\\ 
\hline
\end{tabular}
}

}
\vspace{2mm}
\caption{Prompt template ablations in Nav-GRPO.}
\label{ablation:grpo}
\end{table}

\begin{table}[H]
\centering
\resizebox{\textwidth}{!}{
{
\renewcommand{\arraystretch}{1.3}
\setlength{\tabcolsep}{5pt}
\begin{tabular}{r||ccc||ccc||ccc||ccc||ccc||ccc}
\hline \thickhline

 \rowcolor[HTML]{f8f9fa} \multicolumn{1}{c||}{Reward}& \multicolumn{3}{c||}{Overall} & \multicolumn{3}{c||}{Ins-ImgNav} & \multicolumn{3}{c||}{ImgNav} & \multicolumn{3}{c||}{PointNav} & \multicolumn{3}{c||}{ObjNav} & \multicolumn{3}{c}{VLN}\\
\rowcolor[HTML]{f8f9fa} \multicolumn{1}{c||}{Type} & SR& SPL & OSR & SR& SPL & OSR & SR& SPL & OSR & SR& SPL & OSR & SR& SPL & OSR & SR& SPL & OSR\\
\hline
\hline
Strict & 16.20 & 11.74 & 25.40 & 26.21 & 19.75 & 31.05 & 17.36 & \textbf{14.77} & 20.25 & \bf{22.31} & \bf{16.72} & \bf{26.29} & 43.85 & 31.32 & 60.66 & 28.57 & 24.85 & 45.71\\
Loose & 15.40 & 10.97 & \textbf{26.00} & 27.42 & 18.87 & 31.45 & 18.18 & 13.43 & 21.07 & 19.92 & 13.89 & 22.31 & 43.03 & 33.64 & \textbf{63.11} & 31.43 & 27.33 & 45.71\\ 
Stepped & \textbf{17.00} & \textbf{12.04} & 25.00 & \textbf{29.44} & \textbf{20.73} & \textbf{33.47} & \textbf{19.42} & 14.27 & \textbf{22.73} & 21.51 & 16.37 & 24.70 & \textbf{48.77} & \textbf{35.63} & 62.70 & \textbf{34.29} & \textbf{29.14} & \textbf{48.57}\\
\hline
\end{tabular}
}

}
\vspace{2mm}
\caption{Reward design in Nav-GRPO. }
\label{ablation:reward}
\end{table}

\begin{table}[H]
\centering
\resizebox{\textwidth}{!}{
{
\renewcommand{\arraystretch}{1.3}
\setlength{\tabcolsep}{5pt}
\begin{tabular}{>{\raggedleft\arraybackslash}p{1.7cm}||ccc||ccc||ccc||ccc||ccc||ccc}
\hline \thickhline

 \rowcolor[HTML]{f8f9fa} \multicolumn{1}{c||}{Thinking}& \multicolumn{3}{c||}{Overall} & \multicolumn{3}{c||}{Ins-ImgNav} & \multicolumn{3}{c||}{ImgNav} & \multicolumn{3}{c||}{PointNav} & \multicolumn{3}{c||}{ObjNav} & \multicolumn{3}{c}{VLN}\\
\rowcolor[HTML]{f8f9fa} \multicolumn{1}{c||}{Frequency} & SR& SPL & OSR & SR& SPL & OSR & SR& SPL & OSR & SR& SPL & OSR & SR& SPL & OSR & SR& SPL & OSR\\
\hline
\hline
per 10 steps& 17.00 & 11.51 & 27.60 & 25.81 & 17.80 & 32.66 & 19.83 & 12.95 & 23.14 & 23.51 & \textbf{16.14} & 26.69 & 43.44 & 31.00 & 63.93 & 28.57 & 27.69 & 40.00\\
per 20 steps& \textbf{19.40} & \textbf{13.77} & \textbf{29.40} & \textbf{30.24} & \textbf{20.77} & \textbf{35.48} & \textbf{23.97} & \textbf{17.49} & \textbf{27.27} & 23.51 & 14.35 & 27.89 & \textbf{49.18} & \textbf{37.79} & \textbf{67.21} & \textbf{37.14} & \textbf{33.56} & 42.86\\ 
per 40 steps& 18.80 & 12.62 & 28.20 & 27.42 & 18.46 & 33.87 & 17.77 & 12.00 & 22.73 & \textbf{25.50} & 15.37 & \textbf{29.08} & 48.77 & 34.12 & 64.34 & 34.29 & 26.49 & \textbf{45.71}\\
\hline
\end{tabular}
}

}
\vspace{2mm}
\caption{Thinking frequency ablations.}
\label{ablation:thinking}
\end{table}
\begin{table}[H]
\centering
\resizebox{0.7\textwidth}{!}{
{
\renewcommand{\arraystretch}{1.35}
\setlength{\tabcolsep}{5pt}
\begin{tabular}{r||ccc||ccc||ccc}
\hline \thickhline

 \rowcolor[HTML]{f8f9fa} \multicolumn{1}{c||}{}& \multicolumn{3}{c||}{Easy Instructions} & \multicolumn{3}{c||}{Medium Instructions} & \multicolumn{3}{c}{Hard Instructions} \\
\rowcolor[HTML]{f8f9fa} \multicolumn{1}{c||}{\multirow{-2}{*}{Methods}} & SR& SPL & OSR & SR& SPL & OSR & SR& SPL & OSR\\
\hline
\hline
\rowcolor[HTML]{f8f9fa} \multicolumn{10}{l}{\emph{LVLM as Agent}} \\
Qwen-VL\cite{bai2025qwen2} & 0.00 & 0.00 & 0.00 & 0.00 & 0.00 & 2.96 & 0.00 & 0.00 & 3.29\\
Video-LLaVA~\cite{lin2024video} & 2.23 & 1.25 & 2.23 & 0.00 & 0.00 & 4.14 & 0.00 & 0.00 & 5.26\\
LLaVA-NeXT~\cite{zhang2024llavanextvideo} & 0.56 & 0.49 & 0.56 & 0.00 & 0.00 & 2.37 & 0.00 & 0.00 & 4.61\\
\hline 
\rowcolor[HTML]{f8f9fa} \multicolumn{10}{l}{\emph{Methods for DE}} \\
NaviLLM*~\cite{zheng2024towards} & 2.79 & 2.79 & 2.79 & 0.59 & 0.59 & 5.33 & 0.00 & 0.00 & 4.61\\
NavGPT-2*~\cite{zhou2024navgpt2} & 4.47 & 2.91 & 4.47 & 1.18 & 0.92 & 5.92 & 0.00 & 0.00 & 5.26\\
\hline 
\rowcolor[HTML]{f8f9fa} \multicolumn{10}{l}{\emph{Methods for CE}} \\
NaVid~\cite{zhang2024navid} & 11.17 & 8.89 & 11.17 & 4.14 & 2.98 & 11.24 & 1.32 & 0.51 & 11.84\\
Uni-NaVid~\cite{zhang2025uninavid} & 14.53 & 9.81 & 14.53 & 7.10 & 5.01 & 17.16 & 3.29 & 1.93 & 21.71\\
NaVid\textdagger~\cite{zhang2024navid} & 16.76 & 13.52 & 16.76 & 7.10 & 6.28 & 13.02 & 1.32 & 0.77 & 11.18\\
Uni-NaVid\textdagger~\cite{zhang2025uninavid} & 14.53 & 10.25 & 14.53 & 9.47 & 6.25 & 18.93 & 2.63 & 1.42 & 20.39\\
\hline
OctoNav-R1~(ours) & \textbf{28.49} & \textbf{20.61} & \textbf{28.49} & \textbf{20.71} & \textbf{14.98} & \textbf{32.54} & \textbf{7.24} & \textbf{4.38} & \textbf{26.97}\\
\hline
\end{tabular}
}

}
\vspace{2mm}
\caption{More Comparison with previous methods.}
\label{difficulty}
\end{table}

\subsection{Ablation Study}
\label{append:abla}

We present more detailed ablation results in Tab.~\ref{ablation:grpo}, Tab.~\ref{ablation:reward} and Tab.~\ref{ablation:thinking}.

\subsection{More Comparison Results}
We present more comparison results in Tab.\ref{difficulty}. Specifically, We divide the test instructions into three types according to the number of tasks. Each easy instruction contains only one task. Each medium instruction contains two tasks. A hard instruction contains at least three tasks. As shown in Tab.\ref{difficulty}, OctoNav-R1 achieves the state-of-the-art performance on all types of instructions. Particularly, the SR for hard instructions in improved from $2.63\%$ to $7.24\%$, demonstrating OctoNav-R1's superiority in completing complex instructions.

\section{Appendix: Limitations and Broader Impacts}
\label{append:limitation}
\noindent\textbf{Limitations.}
We find that VLMs sometimes generate hallucinations, leading to degraded navigation performance. Studying how to reduce such hallucinations in VLM research field is also beneficial for embodied navigation.
Considering the cost of the TBA paradigm, we adjust OctoNav-R1 to the mode of thinking at fixed frequency. Overall, the performance is not sensitive to the frequency. However, we believe that when and where to think is a valuable research topic, \eg, slow-fast thinking collaboration and scene-aware adaptive thinking, which we leave for further work.

\noindent\textbf{Broader Impacts.}
Navigation errors by robots in real-world scenarios may lead to safety hazards or property losses. We hope research on enhancing the safety and reliability of robots in real-world environments can facilitate the practical deployment of embodied navigation applications. Moreover, beyond the navigation agents moving on the ground, we hope our work also provides insights for researchers in the drone and UAV literature.


\end{document}